\documentclass[5p]{elsarticle}
\usepackage{lineno,hyperref}
\usepackage{comment}
\usepackage{tikz}
\modulolinenumbers[5]

\journal{Cognition}

\biboptions{authoryear} % APA style citation

\begin{document}

\begin{frontmatter}

\title{The cognitive roots of regularization in language}

\author[mymainaddress,mysecondaryaddress]{Vanessa Ferdinand\corref{mycorrespondingauthor}}
\cortext[mycorrespondingauthor]{Corresponding author at: Santa Fe Institute, 1399 Hyde Park Road, Santa Fe, NM 87501, United States.  Tel.: +1 505 946-2735.}
\ead{vanessa@santafe.edu}
\author[mysecondaryaddress]{Simon Kirby}
\author[mysecondaryaddress]{Kenny Smith}
\address[mymainaddress]{Santa Fe Institute, United States}
\address[mysecondaryaddress]{Centre for Language Evolution, \\ University of Edinburgh, United Kingdom}

\begin{abstract}

Regularization occurs when the output a learner produces is less variable than the linguistic data they observed. In an artificial language learning experiment, we show that there exist at least two independent sources of regularization bias in cognition: a domain-general source based on cognitive load and a domain-specific source triggered by linguistic stimuli.  Both of these factors modulate how frequency information is encoded and produced, but only the production-side modulations result in regularization (i.e. cause learners to eliminate variation from the observed input).  We formalize the definition of regularization as the reduction of entropy and find that entropy measures are better at identifying regularization behavior than frequency-based analyses.  Using our experimental data and a model of cultural transmission, we generate predictions for the amount of regularity that would develop in each experimental condition if the artificial language were transmitted over several generations of learners.  Here we find that the effect of cognitive constraints can become more complex when put into the context of cultural evolution: although learning biases certainly carry information about the course of language evolution, we should not expect a one-to-one correspondence between the micro-level processes that regularize linguistic datasets and the macro-level evolution of linguistic regularity.

\end{abstract} 


\end{frontmatter}

%%%%%%%%%%%%%%%%%%%%%%%%%%%%%%%%%%%%%%%%%%%%%%%%%

\section{Introduction} \label{sec:Intro}

Languages evolve as they pass from one mind to another.  Immersed in a world of infinite variation, our cognitive architecture constrains what we can perceive, process, and produce.  Cognitive constraints, such as learning biases, shape languages as they evolve and can help to explain the structure of language \citep{bever1970cognitive,slobin1973cognitive,newport1988constraints,newport2016statistical,Chater2008,christiansen2016now,Culbertson2012,Kirby2014iterated}.  Early on, debate over the nature of these biases was polarized: Chomsky's nativism program explained linguistic structure as the product of a language-specific acquisition device \citep{Chomsky1957} while behaviorists claimed general-purpose learning mechanisms, such as reinforcement learning, could explain language acquisition \citep{skinner1957verbal}.  Recent experimental research has found domain-general learning mechanisms underpin many aspects of language learning \citep{saffran2007domain}, such as the statistical learning involved in word segmentation by infants \citep{Saffran1996} and how memory constraints modulate learners' productions of probabilistic variation in language \citep{Chang2009}.  
However, it is likely that a mixture of domain-general and domain-specific mechanisms are involved in language learning \citep[e.g.][]{pearl2009domain,culbertson2016simplicity}. 

This paper offers a first attempt to quantify the relative contribution of domain-general and domain-specific learning mechanisms to linguistic regularization behavior.
Regularization is a well-documented process by which learners impose structure on data by reducing the amount of variation in that data.  When language learners encounter linguistic elements in free variation, such as two realizations of a particular phoneme, two synonyms for one meaning, or two possible word orders for constructing a clause, they tend to reduce that free variation by either eliminating one of the variants, or conditioning their variant use on some aspect of the context (e.g. on the adjacent linguistic context). 
Natural languages rarely exhibit free (i.e. unconditioned) variation \citep{givon1985function} and the regularization behavior of language learners is likely to be the cause.
Regularization has been documented extensively in natural language use and in the laboratory.  In natural language, regularization occurs 
in children's acquisition of language 
\citep{Berko1958,Marcus1992,Singleton2004,Smith2007mam},
during the formation of creole languages from highly variable pidgin languages \citep{Bickerton1981,Sankoff1979,Degraff1999,Lumsden1999,Meyerhoff2000,Becker2003},
during the formation of new signed languages 
\citep{Senghas1997,Senghas2000,Senghas2001}, 
and in historical trends of language change 
\citep{Schilling1994,Lieberman2007,vanTrijp2013}.
In the laboratory, regularization has been studied in depth through artificial language learning experiments with children
\citep{HudsonKam2005,HudsonKam2009,Wonnacott2011a,culbertson2015harmonic} 
and adults \citep{Wonnacott2005,HudsonKam2005,HudsonKam2009,Reali2009,Smith2010a,Perfors2012,Culbertson2012,feher2016structural,smith2017language}.  Here we focus on regularization of lexical variation by adult learners in an artificial language learning paradigm.   Future research should explore whether our results generalize to regularization by child learners.

Behavioral experiments offer special insight into the regularization process, because they allow researchers to present participants with controlled linguistic variation, precisely measure the way participants transform that variation, and test hypotheses about what causes participants to alter patterns of variation.  For example, \citet{HudsonKam2009} investigated the regularization of pseudo-determiners in an artificial language learning experiment.  In Experiment 1, adult participants were trained on a language that consisted of several verbs, several nouns (divided into 2 noun classes), 2 \emph{main} determiners (one for each noun class), and zero to 16 \emph{noise} determiners (which could occur with any noun).  In the training language, each noun occurred with its main determiner on 60\% of exposures; the remaining exposures were equally divided across the noise determiners.  In the testing phase, participants described scenes using the language they had learned.  When participants encountered only two noise determiners during training, they regularized slightly by producing the main determiners with 70\% of the nouns, rather than the 60\% they observed in the training language.  Regularization increased with the number of noise determiners, reaching its highest level with 16 noise determiners, where the main determiners were produced with nearly 90\% of the nouns.  In Experiment 2, Hudson Kam \& Newport showed that adult participants regularize the same artificial language less when the noise determiners are conditioned on particular nouns in a more predictable and consistent way.

These results are consistent with Newport's Less-is-More hypothesis.  Originally conceived as an explanation for why children regularize more than adults \citep{Newport1990}, it states that learners with limited memory capacity may regularize inconsistent input because they have difficulty storing and retrieving forms that are lower in frequency or used less consistently.  Regularization behavior varies considerably between children and adults (see e.g. \citealp{HudsonKam2009}, Experiment 3).  However, regularization due to memory limitations may also apply to adults, albeit to a lesser degree \citep{Chang2009}.
Overall, the Less-is-More hypothesis constitutes a domain-general account of linguistic regularization in terms of cognitive constraints on memory encoding and retrieval.  If this hypothesis describes a truly domain-general effect, we should expect to see the same kind of regularization behavior in non-linguistic domains.

\citet{Gardner1957} conducted a frequency prediction experiment in which adult participants had to predict which of several lights would flash in any given trial.  When participants observed two lights flashing at random in a 60:40 ratio (light A flashed 60\% of the time and light B flashed 40\% of the time), they \emph{probability matched} this ratio in their predictions, meaning that about 60\% of their guesses were that light A would flash next and about 40\% of their guesses were on light B.  They also probability matched when observing a 70:30 ratio.  However, when participants were trained on three lights (four ratios were tested: 70:15:15, 70:20:10, 60:20:20, and 60:30:10), they regularized by over-predicting the most frequent light and under-predicting the less frequent lights, which is similar to the behavior of Hudson Kam \& Newport's (2009) participants.  In another experiment, \citet{Kareev1997} report an effect of individual differences in working memory capacity (as determined by a digit-span test) on participants' perception of the correlation of two probabilistic variables. Participants with lower memory capacity overestimated the most common variant, whereas participants with higher capacity did not.  Similarly, \citet{Dougherty2003} show that participants with lower working memory were less likely to consider alternative choices in an eight-item prediction task and were also less likely to consider the low-frequency alternatives than participants with higher working memory.  Each of these cases can be identified as regularization where the higher-frequency variants are over-represented in participants' behavior.

There is therefore strong evidence for the existence of domain-general drivers of regularization, but the extent to which they account for the level of regularity that we observe in language is not clear.  This is because domain-specific learning mechanisms may play a role on their own, or interact with general mechanisms. For example, \citet{Perfors2012} presented seven carefully controlled manipulations of cognitive load during the encoding stage of an artificial language learning task and found no effect on regularization behavior.  This suggests that the Less-is-More Hypothesis may apply more to retrieval than to storage, and that the effects of working memory found in the non-linguistic experiments of \citet{Kareev1997} and \citet{Dougherty2003} may not operate as strongly in language learning.  Furthermore, \citet{Reali2009} show an effect of domain on regularization behavior: participants reduce variation when learning about words but increase variation when learning about coin flips.  However, cognitive load was lower in the coin flipping condition (one coin was flipped, whereas 6 objects were named), so it is unclear whether the higher cognitive load or linguistic domain caused participants to regularize in the word learning task.

In the following, we present a two-by-two experimental design that manipulates cognitive load (following \citealp{HudsonKam2009} and \citealp{Gardner1957}) and task domain (directly comparing regularization in linguistic and non-linguistic domains).  To manipulate cognitive load we vary the number of stimuli a learner must track concurrently.  We manipulate task domain by manipulating the type of stimuli the learner must track: objects being named with words (linguistic domain) or marbles being drawn from containers (non-linguistic domain).  Our method is closely based on the artificial language learning experiment in \citet{Reali2009} and our high load linguistic condition replicates their Experiment 1.

Although we compare regularization behavior in one particular linguistic task (word learning) to one particular non-linguistic task (marble drawing), any differences in regularization behavior revealed by this comparison constitute an existence proof for general and language-specific drivers of regularization behavior.  
Little is known about how regularization behavior compares across different levels of language and the only systematic study of this to date (comparing morphology to word order) reports no global difference in regularization behavior across these two levels \citep{Saldana2017}.
Our two-by-two design can easily be extended to various linguistic tasks at different levels of language (e.g. phonology, morphology, and word order variation) and appropriately matched non-linguistic tasks (coin flipping, flashing light prediction, etc.) to determine the generalizability of the present results.

Based on the work reviewed above, we predict that regularization behavior will increase when cognitive load is raised.  We also predict that regularization behavior will increase when the task is presented with linguistic stimuli.  However, we have no clear prediction about the existence of an interaction between domain and cognitive load, or the relative amount of variation that will be removed from the data due to load or domain.
Knowing the relative contribution of domain-general and domain-specific biases to structure in language is important because it tells us how much we can ground our theories of language learning in general mechanisms of memory and statistical learning.

In order to address these questions, we need a principled measure of regularization that is comparable across different distributions of variation and stimuli domains.  In Section \ref{sec:entropy}, we provide this measure by formalizing the definition of regularization as the reduction of entropy in a data set.  Readers may skip this section if they are willing to accept the following statement: the amount of variation a participant regularizes is equal to the drop in entropy of their productions relative to their observations.
In Section 3, we present the experimental method and design.
In Section 4, we present the main result (both cognitive load and linguistic stimuli elicit regularization), followed by three supporting analyses that explore regularization behavior in greater depth.
In Section 5, we use our empirical data to investigate the evolution of regularity as learners' biases are repeatedly applied under a model of cultural transmission.  This gives us a sense of how predictive known regularization biases can be for the level of regularity found in culturally-transmitted behaviors, such as languages.

%%%%%%%%%%%%%%%%%%%%%%%%%%%%%%%%%%%%%%%%%%%%%%%%%

\section{Defining and quantifying regularization} \label{sec:entropy}

In the existing literature, regularization is described as the elimination or reduction of free variation.  Therefore, we will define regularization in terms of this lost variation and quantify it as the amount of variation that was lost from learners' productions when compared to the data the learners observed.  The amount of variation in any data set can be quantified by the information-theoretic notion of entropy \citep[e.g.][]{Cover1991} and a growing number of studies are using entropy measures to analyze regularization behavior \citep[e.g.][]{Smith2010a, Perfors2012, fedzechkina2014communicative, Ferdinand2015, cuskley2015adoption, perfors2016adult, feher2016structural, smith2017language,Saldana2017,samara2017acquiring}.

The variation in a distribution of items, such as linguistic variants, can be quantified by \emph{Shannon entropy} \citep{Shannon1948}:

\begin{equation}
H(V) = -\sum_{v_i \in V} p(v_i) \log_2 p(v_i)
\end{equation}
\label{eq:ent}

\noindent
where $V$ is the set of linguistic variants in question,
$p(V)$ is the probability distribution over those variants,
and $p(v_i)$ is the probability of $i$th variant in that set.
For example, take the probability distribution over the 4 determiners used in the ``2 noise determiner" condition of \citet{HudsonKam2009}'s artificial language learning experiment: $p(V) = \{0.3, 0.3, 0.2, 0.2\}$.  The Shannon entropy of this distribution is 1.97 bits.
Imagine a participant who was trained on this language and on testing produced the distribution $p(V') = \{0.7, 0.1, 0.1, 0.1\}$. The Shannon entropy of $p(V')$ is 1.36 bits and the change in variation is -0.61 bits.  This means that 0.61 bits of variation among determiners was regularized (i.e. removed) by the participant.  Or, more intuitively, $\frac{0.61}{1.97} \cdot 100 = 31\%$ of the variation in determiners was regularized by the participant.
% unrounded numbers: 1.970951, 1.35678, 0.614171

Variation can also be lost when variants become conditioned on other linguistic variables or contexts.  
For example, each determiner may have a conditional probability $p(v_i|c_j)$ of being produced with a particular noun class $c_j$, such that if one knows the class of the noun, one is better able to predict which determiner a speaker of that language will use with that noun.  The variation in a distribution of items, after a conditioning variable is taken into account, is quantified by \emph{conditional entropy} \citep{Shannon1948}:

\begin{equation}
H(V|C) = -\sum_{c_j \in C} p(c_j) \sum_{v_i \in V} p(v_i|c_j) \log_2 p(v_i|c_j)
\label{eq:condent}
\end{equation}

\noindent
where $V$ is the set of linguistic variants and $C$ is the set of conditioning contexts.
Again, $p(V)$ is the probability distribution over variants,
$p(C)$ is the probability distribution over contexts,
$p(v_i|c_j)$ is the conditional probability of observing the $i$th variant in the $j$th context, 
and $p(c_j)$ is the probability that the $j$th context occurs. 
Given the format of this equation, we can see that the conditional entropy is the sum of the entropy of variants per context, weighted by the probability of each context.
Assume for a moment that the $p(V)$ distribution over determiners is not conditioned on each noun class, meaning that all determiners have the same probabilities regardless of the noun class they are used with, for example: $p(v_i|c_1)$ = \{0.3, 0.3, 0.2, 0.2\} and $p(v_i|c_2)$ = \{0.3, 0.3, 0.2, 0.2\}.  Assume also that any noun has the following probabilities of being in noun class 1 or 2: $p(C) = \{0.6, 0.4\}$.  Let us call this mapping A.  The conditional entropy of mapping A is 1.97 bits, identical to the entropy of the determiners themselves, because the noun class carries no information about which determiner is used.  
We can contrast this with another mapping, mapping B, where determiner use is conditioned on noun class such that $p(v_i|c_1)$ = \{0.5, 0.5, 0.0, 0.0\} and $p(v_i|c_2)$ = \{0.0, 0.0, 0.5, 0.5\}.  Here, the first two determiners in the set are exclusively produced with noun class 1 and the third and fourth determiners are exclusively produced with noun class 2.
The conditional entropy of mapping B is 1.00 bit, while its entropy over determiners remains at 1.97 bits.
If a participant had been trained on a language with mapping A and produced a language with mapping B, then they would have regularized 0.97 bits, or $\frac{0.97}{1.97} \cdot 100 = 49\%$ of the variation in mapping A.

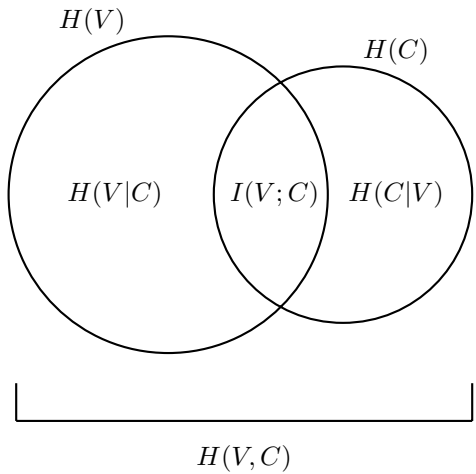
\begin{figure} [t]
\centering
\begin{tikzpicture}[fill=white]
\draw 
[black,thick] (0,0) circle [radius=2.1]
[black,thick] (2.3,0) circle [radius=1.7]
(-1,2.3) node [text=black] {$H(V)$}
(3,1.9) node [text=black] {$H(C)$}
(-0.7,0) node [text=black] {$H(V|C)$}
(3,0) node [text=black] {$H(C|V)$}
(1.4,0) node [text=black] {$I(V;C)$}
(1,-3.5) node [text=black] {$H(V,C)$}
(-2,-3)--(4,-3)
(-2,-3)--(-2,-2.5)
(4,-3)--(4,-2.5);
\end{tikzpicture}
\caption{The relationship between entropy quantities in a mapping between linguistic variants ($V$) and their conditioning contexts ($C$).}
\label{fig:ent_venn}
\end{figure}

Based on these examples, it should be clear that $H(V)$ and $H(V|C)$ are two different kinds of variation that a language can have. $H(V)$ is about the total number of different linguistic forms there are and how often each one is used.  $H(V|C)$ is about how often different linguistic elements occur together.  It is important to note that $H(V)$ and $H(V|C)$, by themselves, do not fully describe the variation in a mapping between linguistic variants and contexts.  Figure \ref{fig:ent_venn} shows the six quantities that are relevant to a complete description of the variation in a linguistic mapping.  The largest quantity, $H(V,C)$, is the total amount of variation in the mapping and is equal to the area covered by the two overlapping circles.  $H(V,C)$ is the joint entropy of the mapping:

\begin{equation}
H(V,C) = -\sum_{v_i \in V} \sum_{c_j \in C} p(v_i,c_j) \log_2 p(v_i,c_j)
\label{eq:joint}
\end{equation}

\noindent
where $p(v_i,c_j)$ is the joint probability of observing the $i$th variant and the $j$th context together.  Looking at Figure \ref{fig:ent_venn}, it is possible to imagine how the joint entropy of the system can increase by moving the two circles away from one another.  As the circles move apart, $V$ and $C$ carry less information about one another.  This has the effect of increasing the two conditional entropy values and reducing the mutual information, $I(C;V)$, between $V$ and $C$.  Mutual information is not a measure of variation, but one of structure: it measures how much uncertainty is reduced in V when C is known, $I(V;C) = H(V) - H(V|C)$, and how much uncertainty is reduced in C when V is known, $I(C;V) = H(C) - H(C|V)$.  Note that $I(V;C) = I(C;V)$.  The five entropy values, on the other hand, are all measures of variation.  Although they each refer to different types of variation, they are related and do constrain one another.  For example, $H(V|C)$ can never be larger than $H(V)$ and $H(C|V)$ can never be larger than $H(C)$.  It is also important to note that $H(V|C) \neq H(C|V)$ (because $H(V)$ and $H(C)$ can take different values).  Regularization can be quantified in terms of these five entropy values and be said to occur when one or more of these values decreases. \\

\indent {\bf Regularization} is the reduction or elimination \\
\indent of entropy in a data set. \\

We define regularization as any reduction to the space in Figure \ref{fig:ent_venn}.  Regularization can occur by eliminating linguistic variants (reducing $H(V)$), eliminating conditioning contexts (reducing $H(C)$), or increasing the degree to which variants and contexts are conditioned on one another (reducing $H(V|C)$ and/or $H(C|V)$).  Joint entropy always decreases when there is a net loss of variation.  Mutual information, on the other hand, does not necessarily change when regularization occurs.  In the following experiment, we construct a stimuli set in which lexical items are variants and the objects they refer to are contexts.  In a matched non-linguistic stimuli set, marbles are variants and the containers they are drawn from are contexts.  The experiment is designed such that $H(V)$, $H(V|C)$, and $H(V,C)$ will always change by the same number of bits when participants regularize and $H(C)$, $H(C|V)$, and $I(V;C)$ can not be changed by participants.

%%%%%%%%%%%%%%%%%%%%%%%%%%%%%%%%%%%%%%%%%%%%%%%%%

\section{Frequency learning experiment} \label{sec:experiment}

In this experiment we manipulate cognitive load and task domain, allowing us to quantify the amount of variation participants regularize due to each source.  Participants observe an input mapping among stimuli and then produce behavior from which an output mapping is extracted.  Finally, they estimate the frequencies of the input stimuli and these estimates are compared to their output behavior.

\subsection{Participants} 

573 participants were recruited via Amazon's Mechanical Turk crowdsourcing platform and completed our experiment online. Informed consent was obtained for experimentation.  Participant location was restricted to the USA and verified by a post-hoc check of participant IP address location.  61 participants were excluded on the basis of the following criteria: failing an Ishihara color vision test (15), self-reporting the use of a pen or pencil during the task in an exit questionnaire (10), not reporting their sex or age (6), self-reporting an age below 18 (1), or having previously participated in this or any of our related experiments, as determined by their user ID with MTurk (26).  More participants were recruited than necessary with the expectation that some would be excluded by these criteria.  Once the predetermined number of participants per condition was met, the last participants were excluded (3).  All participants (included and excluded) received the full monetary reward for participating in this task, which was 0.10 USD in the one-item conditions (\emph{marbles1} and \emph{words1}) and 0.60 USD in the six-item conditions (\emph{marbles6} and \emph{words6}).\footnote{Data collection began in 2012, when this was standard reimbursement for participants recruited through MTurk.  In current practice, standard reimbursement is US federal minimum wage.}  
The average time taken to complete the one-item conditions was 3 minutes and 50 seconds, with a standard deviation of 1 minute and 27 seconds.  Average time to complete the six-item conditions was 11 minutes and 32 seconds, with a standard deviation of 2 minutes and 6 seconds.
% this is duration from HIT accept to HIT submit.
Of the final 512 participants, 274 reported female, 238 reported male, and the mean age was 33.7 years (min = 18, max = 72) with a standard deviation of 11.3 years.

\subsection{Materials and Stimuli}

The experiment was coded up as a Java applet that ran in the participant's web browser in a 600x800-pixel field. Photographs of 6 different containers (a bucket, bowl, jar, basket, box, and pouch) and computer-generated images of marbles in 12 different colors (blue, orange, red, teal, pink, olive, lime, purple, black, yellow, grey, and brown) served as non-linguistic stimuli.  Photographs of 6 different novel objects (resembling mechanical gadgets) and 12 different nonsense words (\emph{buv}, \emph{kal}, \emph{dap}, \emph{mig}, \emph{pon}, \emph{fud}, \emph{vit}, \emph{lem}, \emph{seb}, \emph{nuk}, \emph{gos}, \emph{tef}) served as linguistic stimuli.  Stimuli were chosen to have similar visual complexity across domain (determined by gzip complexity and area of stimuli). 
Marbles and words were organized into fixed pairs that maximized distinctiveness between the stimuli in the pair.  The stimuli lists above appear in order of these pairings (blue and orange were paired, \emph{buv} and \emph{kal} were paired, etc.).  Marble colors were paired to differ in hue and brightness.  Within-pair hue differences were greater than $120^{\circ}$ (i.e. chosen from approximately opposite sides of the color wheel) and within-pair brightness differences were greater than 20\%.  Words were paired to be contrastive.  Within-pair words utilized different letters and vowels and within-pair consonants differed by place of articulation.  These stimuli are closely based on the word stimuli used in \citet{Reali2009} and selected to not look or sound like existing words when pronounced by an American English speaker.  Words were presented visually and were not accompanied by auditory stimuli.

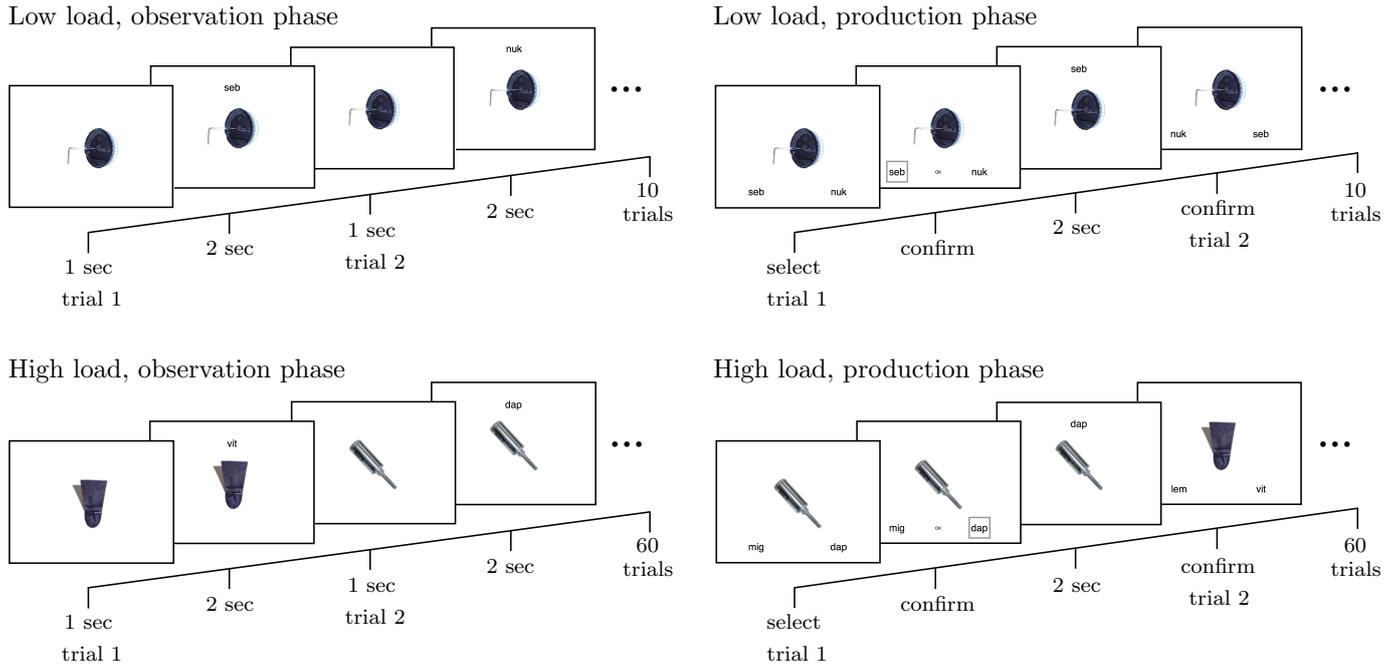
\begin{figure*} [t] \centering
\centering
\includegraphics[width=180mm]{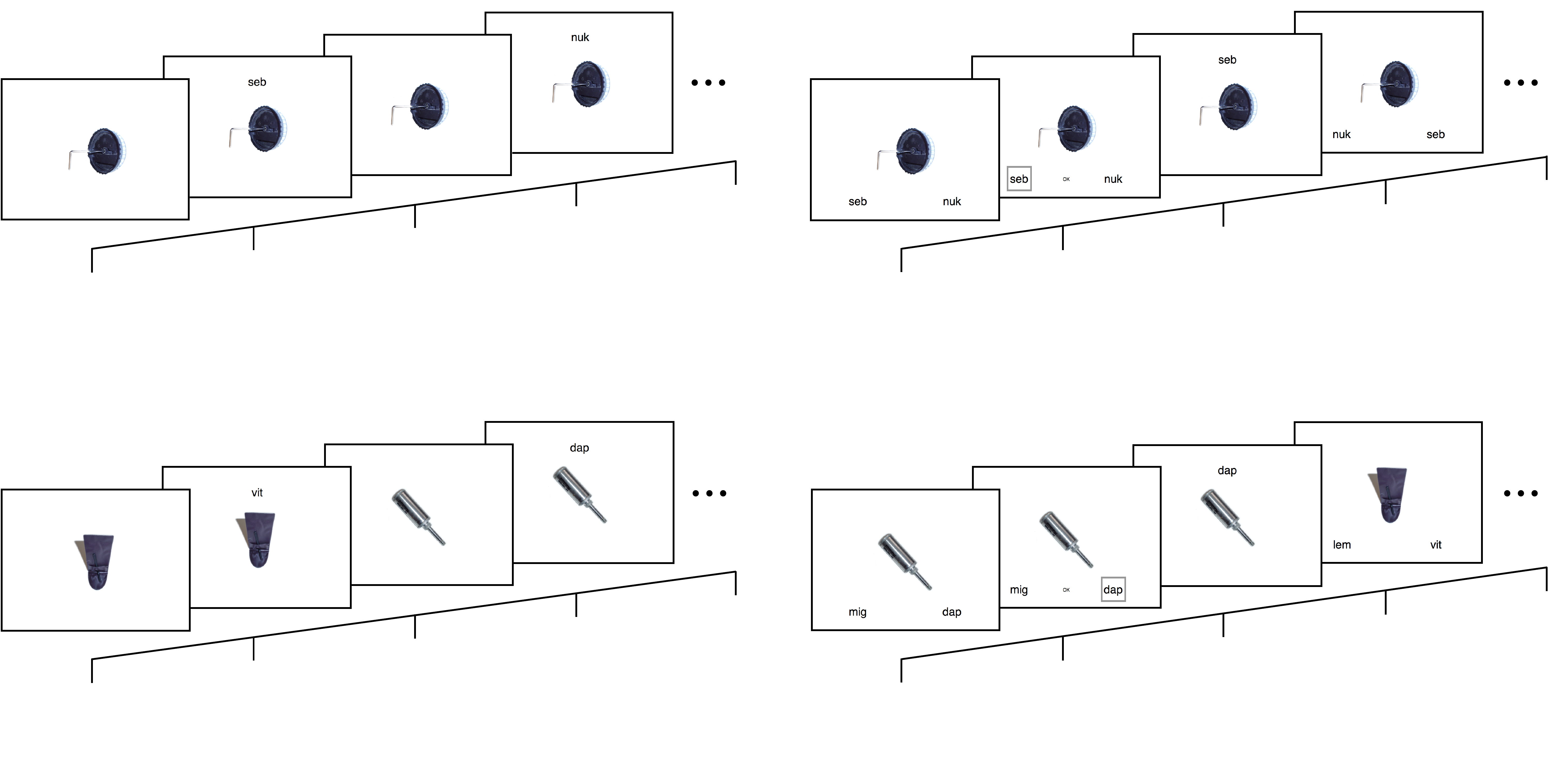}
\put(-512,246){Low load, observation phase} % top
\put(-248,246){Low load, production phase}
\put(-512,113){High load, observation phase} % bottom
\put(-248,113){High load, production phase}
\footnotesize
\put(-492,140){trial 1} % top
\put(-386,154){trial 2}
\put(-277,181){10} \put(-282,172){trials}
\put(-228,140){trial 1} 
\put(-70,162){trial 2}
\put(-12,181){10} \put(-17,172){trials}
\put(-492,6){trial 1} % bottom schema
\put(-386,20){trial 2}
\put(-277,47){60} \put(-282,38){trials}
\put(-228,6){trial 1}
\put(-70,27){trial 2}
\put(-12,47){60} \put(-17,38){trials}
\footnotesize
\put(-491,152){1 sec} % top
\put(-438,159){2 sec}
\put(-385,166){1 sec}
\put(-333,173){2 sec}
\put(-228,152){select}
\put(-178,159){confirm}
\put(-121,167){2 sec}
\put(-73,174){confirm}
\put(-491,18){1 sec} % bottom
\put(-438,25){2 sec}
\put(-385,32){1 sec}
\put(-333,39){2 sec}
\put(-228,18){select}
\put(-178,25){confirm}
\put(-121,32){2 sec}
\put(-73,39){confirm}
\caption{Schema of the experiment's observation and production phases.  \emph{Top:} Low cognitive load condition.  \emph{Bottom:} High cognitive load condition.
Examples shown are the linguistic condition. In the non-linguistic condition, containers are shown in place of the object, and marbles are shown in place of the words.}
\label{fig:schema}
\end{figure*}

\subsection{Conditions and Design} \label{Exp2_conditions}

We use a two-by-two design to investigate the effects of domain and cognitive load in four experimental conditions: \\

\noindent
1) \emph{Non-linguistic single frequency learning (marbles1)} \\
Participants observed two marble colors being drawn from one container at a particular ratio (for example, 5 blue marbles and 5 orange marbles displayed in random order).  Participants were then asked to demonstrate what another several draws from the same container are likely to look like.  They were not asked to predict specific future draws and thus no feedback was given.  Participants observed 10 marble draws and produced 10 marble draws.  
Each participant observed a set of draws in one of six possible ratios:  5:5, 6:4, 7:3, 8:2, 9:1, and 10:0.  These constitute six \emph{input ratio conditions}.  We will refer to the ratio that a participant observed as the \emph{input ratio} and the ratio that the participant produced as the \emph{output ratio}.  There were 32 participants in each input ratio condition, totaling 192 participants in \emph{marbles1}.  Container stimuli were randomized across participants: each participant saw one of the six containers.  Equal numbers of participants saw each container.  Marble pairs were also randomized across participants: each participant saw one of the six marble pairs.  Equal numbers of participants saw each marble pair.  One variant in each pair was randomly assigned to be the majority variant (i.e. have the frequency of 6, 7, 8, 9, or 10).  The full details of the observation and production regimes can be found in Section \ref{sec:procedure} and Figure \ref{fig:schema}. \\

\noindent
2) \emph{Non-linguistic multiple frequency learning (marbles6)} \\
This condition is similar to the \emph{marbles1} condition, with the difference that participants observed and produced 10 draws each from 6 different containers, where each container differed in the ratio of the two marble colors.  Containers, marble pairs, and input ratios were randomly assigned to one another, without replacement, and these assignments were randomized between participants.  Each participant saw all six of the containers, all six of the marble pairs, and all six of the input ratios (the same input ratios as were used in the \emph{marbles1} condition: 5:5, 6:4, 7:3, 8:2, 9:1 and 10:0).  There were 64 participants in this condition, yielding data for 384 (64x6) input ratios. \\

\noindent
3) \emph{Linguistic single frequency learning (words1)} \\
This condition is similar to the \emph{marbles1} condition, differing only by the use of linguistic stimuli (objects and words) instead of the non-linguistic stimuli (containers and marbles) and minimal adaptation of the instructions to the linguistic domain.  Participants observed one object being named with two words at a particular ratio (for example, \emph{buv} 5 times and \emph{kal} 5 times, in random order) and were then asked to name the object like they had observed it being named. 
They were not asked to predict specific future namings and thus no feedback was given.  Participants observed 10 namings and produced 10 namings.  Each of the 6 possible input ratios (same ratios as used in \emph{marbles1}) was observed by 32 participants, totaling 192 participants. \\

\noindent
4) \emph{Linguistic multiple frequency learning (words6)} \\
This condition is similar to the \emph{marbles6} condition, again differing only by the use of linguistic stimuli and minimal adaptation of the instructions to the linguistic domain.  This condition constitutes a replication of the word learning experiment in \citet{Reali2009}, but with different object stimuli, modified word stimuli, and participants who completed the experiment online rather than in the laboratory.  There were 64 participants in this condition, yielding data for 384 (64x6) input ratios.

\subsection{Procedure} \label{sec:procedure}

\begin{figure} [h]
\includegraphics[width=89mm]{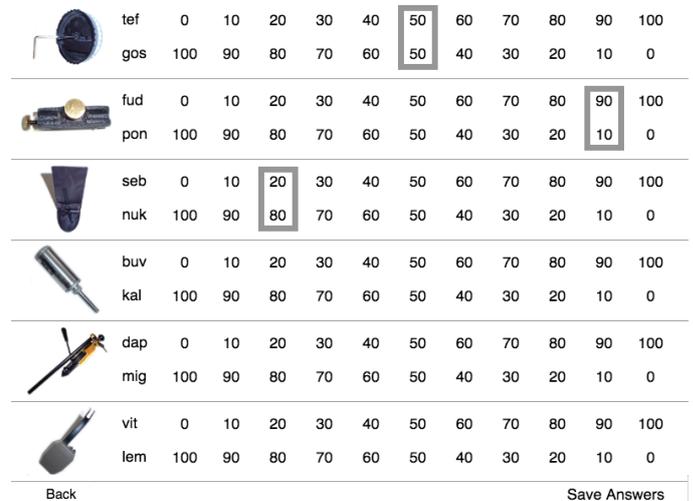}
\caption{Screen shot of the sliders page in the high cognitive load linguistic condition, showing three answers selected.  Participants could change their answers up until ``Save Answers" was clicked.  ``Back" took participants back to the question and instruction about the sliders.  In the low load condition, only one slider was shown.}
\label{fig:sliders}
\end{figure}

The experiment consisted of an observation phase and a production phase.  Figure \ref{fig:schema} shows the structure and timing of the trials.  Participants were not told how many observation or production trials there would be.  In the observation phase, marble/word stimuli were presented in random order.  In the high load conditions, the containers/objects were presented in random order.  In each production trial, the left-right location of the two marbles/words was randomized.  When the participant moused over an answer, a 100x100 pixel box was displayed around the choice.  When clicked, the box remained and an OK button appeared equidistant between the two choices.  Participants could change their answer and clicked OK to confirm their final response.   Their choice was shown over the container/object and the next trial began.  The OK button served to re-center the participant's cursor between trials.

After the production phase, participants were asked to estimate the generating ratio that underlies the input ratio they saw.  This was accomplished by asking them how many marbles of each color were in each container, or how often each word is said for each object in the artificial language.
Participants provided their response with a discrete slider over 11 options of relative percentages (Figure \ref{fig:sliders}).

\subsection{Entropy of the training stimuli set} \label{sec:stim_entropy} \label{sec:stimset_ent}

\begin{table}[t]
\centering
\scalebox{0.8}{
\begin{tabular}{ | l | c | c | c | c | c | c | c | c | c | c | c |  c |}
\hline
& $v_1$ & $v_2$ & $v_3$ & $v_4$ & $v_5$ & $v_6$ & $v_7$ & $v_8$ & $v_9$ & $v_{10}$ & $v_{11}$ & $v_{12}$ \\
\hline
$c_1$ & 5 & 5 & 0 & 0 & 0 & 0 & 0 & 0 & 0 & 0 & 0 & 0  \\
\hline
$c_2$ & 0 & 0 & 6 & 4 & 0 & 0 & 0 & 0 & 0 & 0 & 0 & 0  \\
\hline
$c_3$ & 0 & 0 & 0 & 0 & 7 & 3 & 0 & 0 & 0 & 0 & 0 & 0  \\
\hline
$c_4$ & 0 & 0 & 0 & 0 & 0 & 0 & 8 & 2 & 0 & 0 & 0 & 0  \\
\hline
$c_5$ & 0 & 0 & 0 & 0 & 0 & 0 & 0 & 0 & 9 & 1 & 0 & 0  \\
\hline
$c_6$ & 0 & 0 & 0 & 0 & 0 & 0 & 0 & 0 & 0 & 0 & 10 & 0  \\
\hline
\end{tabular}}
\caption{Co-occurance frequencies among the twelve variants and six contexts in the experimental stimuli set.  Each cell gives the number of times that the participant observed variant{$_i$} along with context{$_j$}.}
\label{tab:freqs}
\end{table}

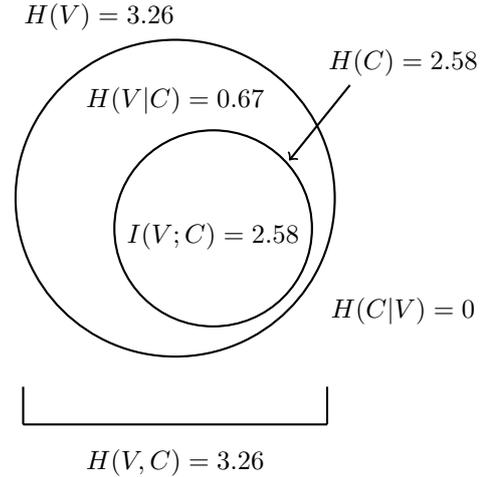
\begin{figure} [t]
\centering
\begin{tikzpicture}[fill=white]
% outline
\draw 
[black,thick] (0,0) circle [radius=2.1]
[black,thick] (0.5,-.4) circle [radius=1.3]
(-1,2.4) node [text=black] {$H(V) = 3.26$}
(3,1.8) node [text=black] {$H(C) = 2.58$}
(0,1.3) node [text=black] {$H(V|C) = 0.67$}
(3,-1.5) node [text=black] {$H(C|V) = 0$}
(0.5,-0.5) node [text=black] {$I(V;C) = 2.58$}
(0,-3.5) node [text=black] {$H(V,C) = 3.26$}
(-2,-3)--(2,-3)
(-2,-3)--(-2,-2.5)
(2,-3)--(2,-2.5);
\draw [<-,thick] (1.5,0.5)--(2.3,1.5);
\end{tikzpicture}
\caption{Entropy of the training stimuli (in bits).  In the linguistic condition, $V$ is the distribution over words and $C$ is the distribution over objects.  In the non-linguistic condition, $V$ is the distribution over marbles and $C$ is the distribution over containers.  Refer back to Section \ref{sec:entropy} for the definition of each quantity.  The experiment is designed so participants can change the size of the outer circle only.}
\label{fig:stim_venn}
\end{figure}

Each participant observes a stimuli set that is composed of co-occurrances between marbles and containers or words and objects.  For the purpose of quantifying the variation in the stimuli sets, we consider the marbles and words to be variants and consider the containers and objects to be contexts.  Table \ref{tab:freqs} shows the co-occurrance frequencies between contexts and variants.  In the high cognitive load conditions, this table describes the complete stimuli set that each participant was trained on in the observation phase.  In the low cognitive load conditions, each participant was trained on only one row from the this table.  Figure \ref{fig:stim_venn} shows the entropy values associated with Table \ref{tab:freqs} and describes the population-level variation in stimuli.  These values are the same across conditions, allowing the direct comparison of mean change in entropy between conditions.

It is important to note that the experimental design prevents participants from changing $H(C)$ because contexts are presented the same number of times in the observation and production phases. $H(C|V)$ cannot be changed either because the only production options are the two variants that were shown per context in the observation phase.  If participants regularize, $H(V)$, $H(V|C)$, and $H(V,C)$ will drop by the same number of bits.

The entropy of the stimuli that one participant observes in the high cognitive load condition is identical to Figure \ref{fig:stim_venn}.  However, the entropy of stimuli in the low cognitive load condition is lower and varies by the input ratio observed: in condition 5:5, 6:4, 7:3, 8:2, 9:1, and 10:0, $H(V) = H(V|C) = H(V,C) = I(V;C) = $ 1 bit, 0.97 bits, 0.88 bits, 0.72 bits, 0.47 bits, and 0 bits, respectively, and $H(C) = H(C|V) = 0$.  

\begin{figure*} [t] \centering
\includegraphics[width=140mm]{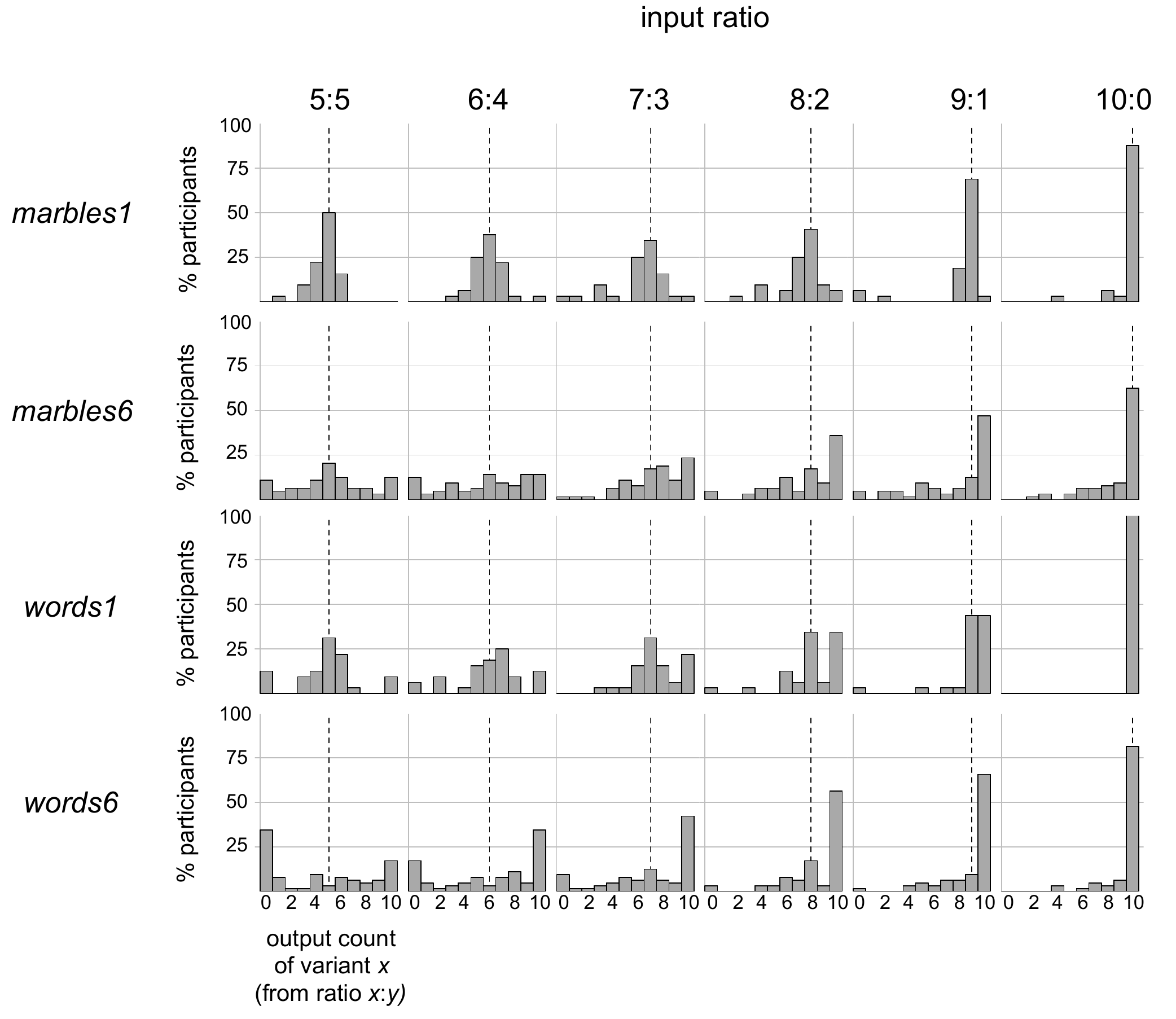}
\caption{Each row shows the results of one experimental condition.  Each column corresponds to one of the six input ratios, ranging from 5:5 (left) to 10:0 (right).
Each pane contains the distribution of output ratios that participants produced in response to one input ratio.
Output ratios are displayed on the x-axis as the number of times a participant produced variant \emph{x} from the input ratio \emph{x:y}, where variant \emph{x} corresponds to whatever marble/word was in the majority during the observation phase.  (In the 5:5 input ratio a random marble/word was coded as variant \emph{x}.)  All input ratios are indicated by a dashed line.}
\label{fig:2x2_6panels}
\end{figure*}

%%%%%%%%%%%%%%%%%%%%%%%%%%%%%%%%%%%%%%%%%%%%%%%%%

\section{Results} \label{sec:results}

First, we describe participant behavior and present the main result: cognitive load and linguistic stimuli both elicit regularization behavior.  Next, three supporting analyses explore regularization behavior in greater depth:  Section \ref{sec:encoding} shows that participants' regularization behavior was due to production biases rather than an encoding bias, Section \ref{sec:individual} analyzes individual differences in regularization during this experiment, and Section \ref{sec:primacy} shows primacy effects help explain why some individuals regularized with the minority variant, rather than the majority variant.

\subsection{Regularization behavior profiles}

Before analyzing the data in terms of its entropy, we first visually inspect how participants changed each input ratio.  In Figure \ref{fig:2x2_6panels}, each panel shows the distribution of ratios that participants produced in response to each input ratio they observed, per experimental condition.

The first row (\emph{marbles1}) shows clear probability matching behavior, where both the mean and mode of participant responses are near the input ratio.  Participants in this condition tended to successfully reproduce their input ratio, with a small amount of error.  The second row (\emph{marbles6}) shows clear regularization behavior.  Participants in this condition have moved distributional mass away from the input ratio and toward the maximally regular ratios, 0:10 and 10:0.  Responses to the 5:5 input ratio seem to be a combination of probability matching behavior (13 participants also produced a 5:5 ratio) and regularization behavior (15 participants produced maximally regular ratios).

The third row (\emph{words1}) shows a mixture of probability matching and regularization behavior for all input ratios.  Roughly half of the participants appear to have probability matched with error rates similar to \emph{marbles1}, and roughly half of the participants appear to have regularized at levels comparable to \emph{marbles6}.  In the 10:0 input condition, none of the participants choose the unseen word on any production trial.  The fourth row (\emph{words6}) shows a similar regularization profile to \emph{marbles6}, but with a more extreme movement of distributional mass to the edges, such that the majority of participants produced maximally regular ratios.  This condition constitutes a successful replication of the first experiment reported in \citet{Reali2009}.

\subsection{Regularization per condition} \label{sec:netreg}

\begin{figure} [t]
\includegraphics[width=90mm]{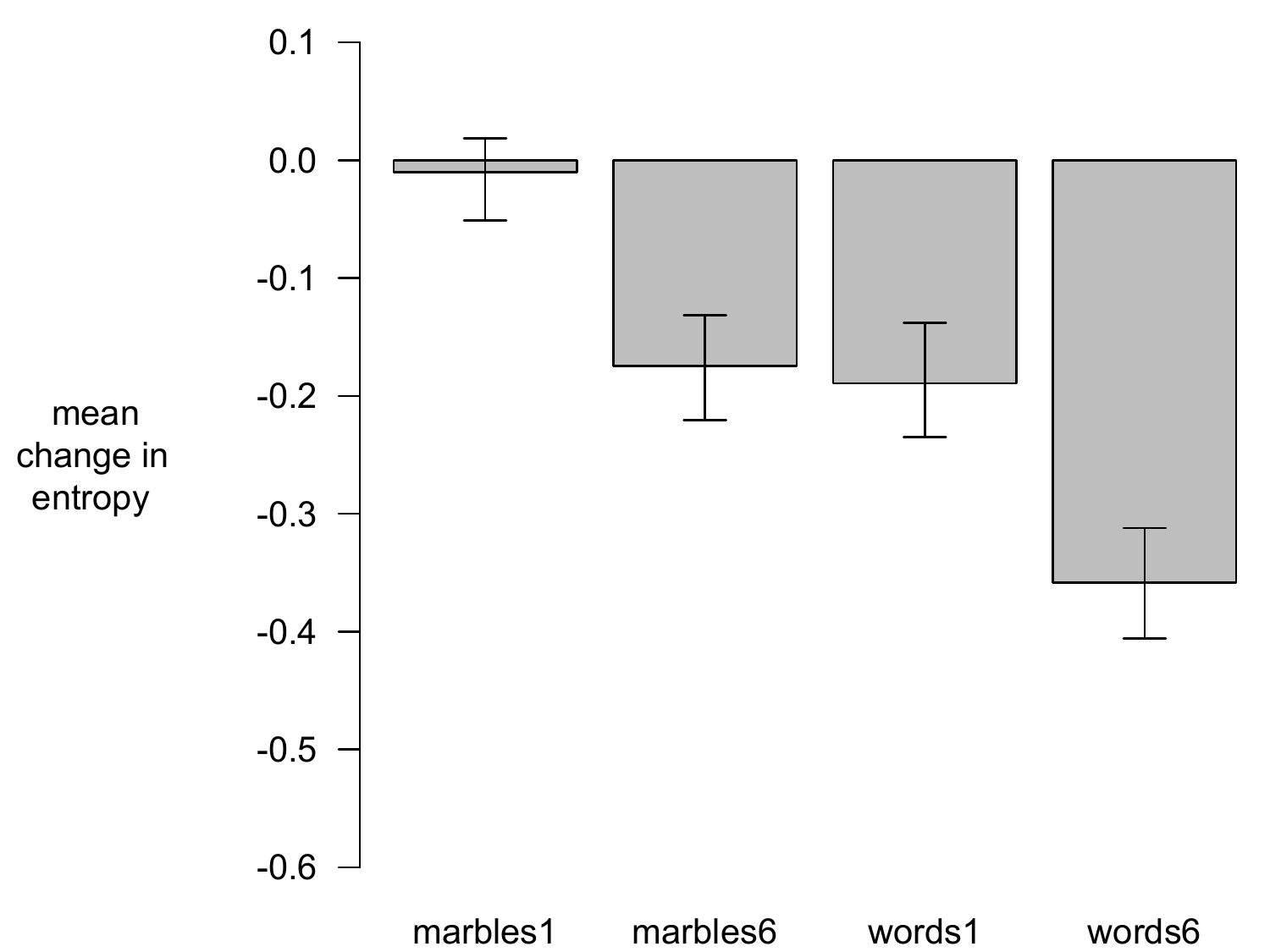}
\put(-121.5,166){***}
\put(-77,166){***}
\put(-32.5,166){***}
\caption{Entropy drops when learners regularize.
Each bar shows the average change in Shannon entropy over all pairs of input-output ratios, per condition.  Stars indicate significant difference from zero.  Error bars indicate the 95\% confidence intervals computed with the bootstrap percentile method \citep{Efron1979}.  A significant drop in entropy means that participants regularized in that condition.  Non-significant differences from zero are obtained when participants probability match.  The lower and upper bounds on mean entropy change for this experiment are $-0.67$ and $+0.33$ bits.}
\label{fig:4bars}
\end{figure}

In this section, we report the differences in regularization behavior within and between the four experimental conditions.  We do this by calculating the change in Shannon entropy for each pair of input-output ratios obtained from participants.  For example, if a participant observes a 5:5 ratio of orange and blue marbles for the jar, and then produces a 6:4 ratio of orange and blue marbles for the jar, the Shannon entropy for that pair of input-output ratios changes by $-0.12$ bits.\footnote{From here onward, whenever we refer to the ``entropy of a ratio" we mean the Shannon entropy of the two variants in ratio $x$:$y$, where the probability distribution over the variants is $p(V) = \{\frac{x}{10},\frac{y}{10}\} $.}  Figure \ref{fig:4bars} shows the mean change in entropy for all input-output ratio pairs per condition.  Negative values mean participants made ratios more regular on average.

To assess the significance of differences in regularization within and between conditions, a linear mixed effects regression analysis was performed using R \citep{R2013} and \emph{lme4} \citep{Bates2013}.  The dependent variable was the change in entropy of the input-output ratios.  Experimental condition was the independent variable.  Participant was entered as a random effect (with random intercepts).  No obvious deviations from normality or homoscedasticity were apparent in the residual plots.

Within-condition changes were assessed by re-leveling the model to obtain the intercept value for each condition.  The intercept equals the condition's mean change in entropy and the regression analysis provides a t-statistic to evaluate whether this mean is significantly different from zero.  Three of the four experimental conditions elicited a significant amount of regularization behavior (Figure \ref{fig:4bars}).  Participants regularized an average of 0.17 bits in \emph{marbles6} ($S.E. = 0.03, t(1152) = -5.53, p < .001$), 0.19 bits in \emph{words1} ($S.E. = 0.03, t(1152) = -6.52, p < .001$), and 0.36 bits in \emph{words6} ($S.E. = 0.03, t(1152) = -11.34, p < .001$).  In \emph{marbles1}, the mean loss of 0.01 bits was not significantly different from zero, which indicates that participants are probability matching in this condition ($S.E. = 0.03, t(1152) = -0.35, p = 0.73$). Overall, participants regularized 26\%, 28\%, and 53\% of the conditional entropy in \emph{marbles6}, \emph{words1}, and \emph{words6}, respectively.

Pairwise comparison of regularization between conditions is also obtained from this re-leveled model.  All pairwise comparisons showed a significant difference in regularization behavior at the $p < .001$ level, except for that between \emph{words1} and \emph{marbles6} ($S.E. = 0.04, t(1152) = 0.34, p = 0.73$).  

\subsection{Domain vs. cognitive load} \label{sec:manips_lmer}

Effects of the experimental manipulations were assessed by constructing a full linear mixed effects model with three independent variables (i.e. fixed effects) and their interaction: domain, cognitive load, and entropy of the input ratio.  The dependent variable was the change in entropy of the input-output ratios.  Participant was entered as a random effect (with random intercepts).  The significance of each fixed effect was determined by likelihood ratio tests, on the full model (described above) against a reduced model which omits the effect in question.  There was a significant effect of domain ($\chi^2(4) = 46.048, p < .001$), cognitive load ($\chi^2(4) = 105.07, p < .001$), and input ratio ($\chi^2(4) = 520.23, p < .001$).  Interactions between fixed effects were also determined by likelihood ratio tests by comparing a reduced model (which omits all interactions) to one which includes the interaction of interest.  Two interactions were found to be significant: cognitive load and input ratio ($\chi^2(1) = 74.695, p < .001$) and domain and input ratio ($\chi^2(1) = 4.4462, p = 0.03$).  The interaction between domain and cognitive load was not significant ($\chi^2(1) = 0.0059, p = 0.94$). 

\begin{figure} [t]
\includegraphics[width=90mm]{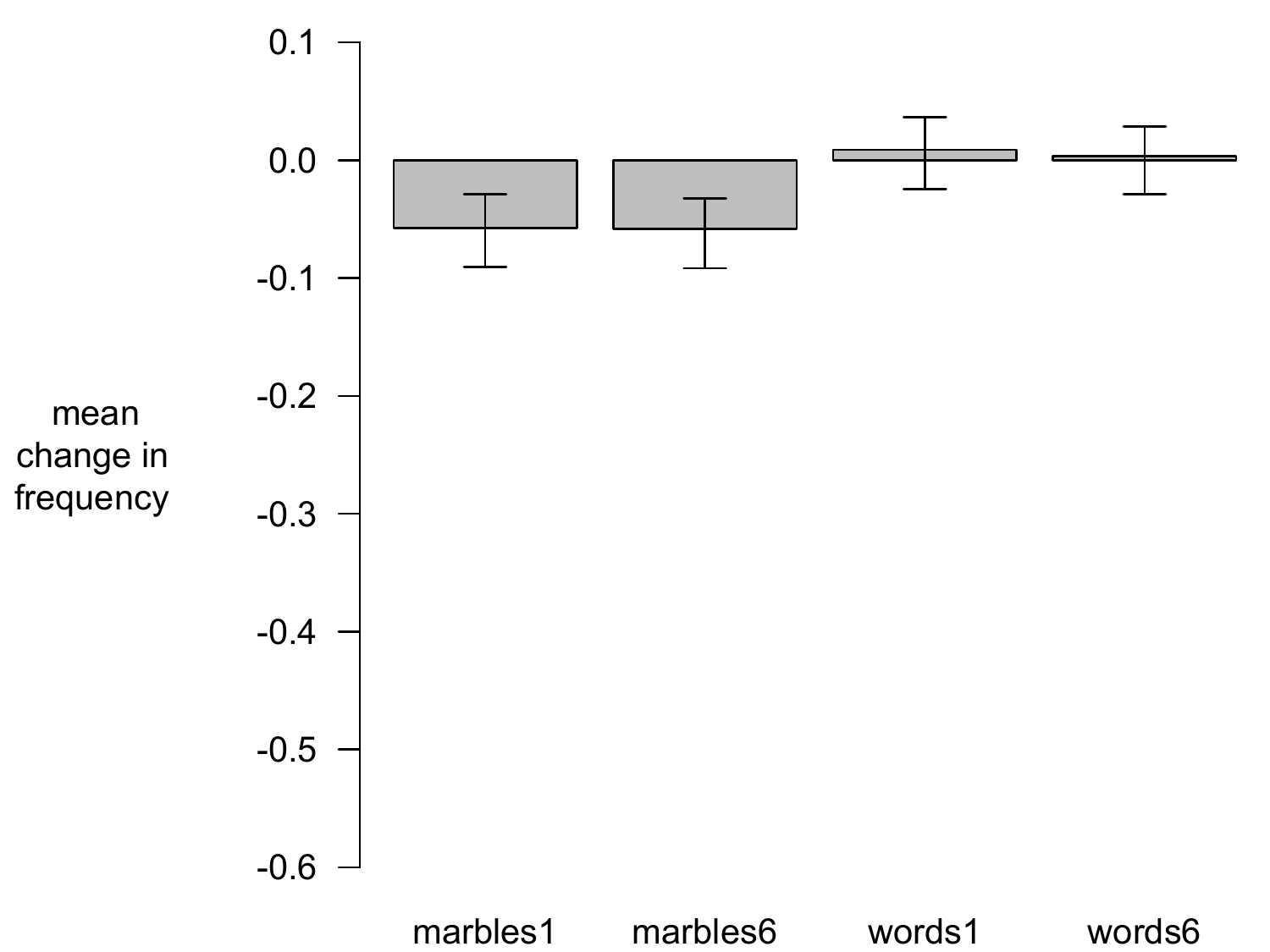}
\put(-166,166){***}
\put(-121.5,166){***}
\caption{Raw changes in frequency fail to capture regularization behavior.  Each bar shows the average difference between the number of times participants observed the majority variant in the training set and the number of times they produced that variant in the testing phase.  Error bars indicate the 95\% confidence intervals computed with the bootstrap percentile method \citep{Efron1979}.  Values significantly higher than zero indicate a population-level trend in over-producing the majority variant.  Values significantly lower than zero indicate a population-level trend in over-producing the minority.
}
\label{fig:maj_4bars}
\end{figure}

Therefore, the best-fit model contained an interaction between domain and input ratio, an interaction between cognitive load and input ratio, but only an additive relationship between domain and cognitive load (loglikelihood $= -278.71$).  A summary of the best-fit model is included in the Appendix, Table \ref{tab:bestfit1}.  The effect of input ratio on entropy change is due to different amounts of regularization being possible under each input ratio (the maximum drop in entropy achievable under the 5:5 through 0:10 ratios are 1, 0.97, 0.88, 0.72, 0.47, and 0 bits, respectively).  As input entropy increases from 0 to 1 bits, output entropy changes by $-0.14$ bits.
This means that participants regularize more when the entropy of the input ratio increases from 0 bits (the 10:0 ratio) to 1 bit (the 5:5 ratio).  The interactions mean that the effect of input entropy on output entropy is greater by $-0.1$ bits when linguistic stimuli are used and greater by $-0.5$ bits when cognitive load is high.  The additive relationship suggests that domain and cognitive load are independent drivers of regularization behavior.

\subsection{Frequency-based analysis of regularization} \label{sec:freq_analysis}

In much of the linguistic regularization literature to date, regularization is measured in terms of stimulus frequency, rather than entropy.  In this section, we repeat the analyses from Section \ref{sec:netreg} and \ref{sec:manips_lmer} with a different dependent variable, \emph{change in frequency of the majority variant} \citep[as in, e.g.][]{HudsonKam2005, Reali2009}, to illustrate the difference between these two approaches.
Figure \ref{fig:maj_4bars} shows the mean change in frequency of the majority variant ($x$ from input ratio $x$:$y$).  For example, if a participant produces a 7:3 ratio in response to a 9:1 input ratio, there is a -0.2 change in majority variant frequency for that pair of input-output ratios.  In the 5:5 input condition, a random variant was encoded as the ``majority" variant.  Positive changes mean participants over-produced the majority variant and negative changes mean participants over-produced the minority variant.  In Figure \ref{fig:maj_4bars} we see that none of the conditions elicit over-production of the majority variant on average, despite the fact that participants in \emph{marbles6}, \emph{words1}, and \emph{words6} are clearly regularizing input ratios (compare to Figure \ref{fig:4bars}).

Applying the analysis in Section \ref{sec:netreg} to the change in majority variant frequency, we find that neither linguistic condition shows a significant change in majority variant frequency (\emph{words1}: $S.E. = 0.02, t(1152) = 0.22, p = 0.82$; \emph{words6}: $S.E. = 0.01, t(1152) = -0.65, p = 0.51$).  However, the frequency-based analysis does reveal something that the entropy-based analysis was unable to capture: a significant over-production of the \emph{minority} variant in the marble-drawing domain, \emph{marbles1} ($S.E. = 0.02, t(1152) = -2.882$, $p=.004$) and \emph{marbles6} ($S.E. = 0.01, t(1152) = -3.269$, $p=.001$).

To determine the effects of the experimental manipulations, we apply the analysis in Section \ref{sec:manips_lmer} to the change in majority variant frequency (and we change the fixed effect \emph{entropy of the input ratio} to \emph{input frequency of the majority variant} in order to match the dependent variable).  We find a significant effect of domain ($\chi^2(4) = 16.391, p = 0.003$) and input frequency ($\chi^2(4) = 14.634, p = 0.006$) on change in majority variant frequency, but no significant effect of cognitive load ($\chi^2(4) = 3.0755, p = 0.55$).  We also find a significant interaction between domain and input frequency ($\chi^2(4) = 6.7741, p = 0.009$).  
Therefore, the best-fit model contains an effect of domain, input frequency, and an interaction between domain and input frequency (loglikelihood $= -77.0$, see Appendix Table \ref{tab:bestfit_freq}).

In summary, the frequency analysis fails to capture the effect of cognitive load on regularization behavior and fails to capture the fact that participants are eliminating variation in the linguistic domain.  The reason mean change in frequency is not different than zero in the linguistic domain is because participants sometimes regularized with the majority variant and other times regularized with the minority variant, in a way that tends to cause frequency changes to average out to zero.  However, as is clear from the raw data, it would be incorrect to conclude that participants are probability matching in the linguistic domain.

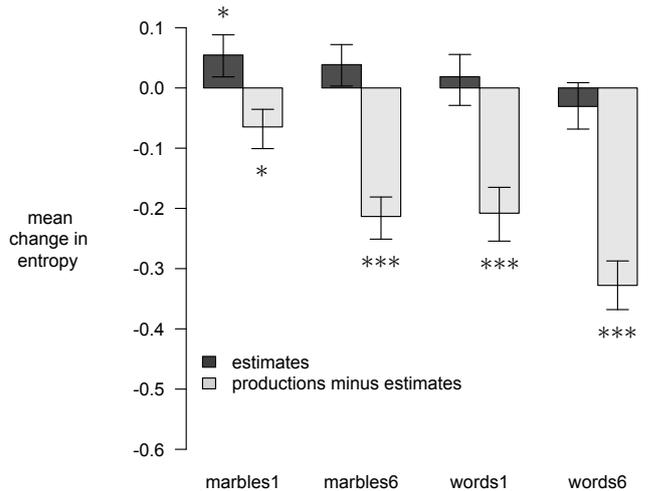
\begin{figure} [t]
\includegraphics[width=86mm]{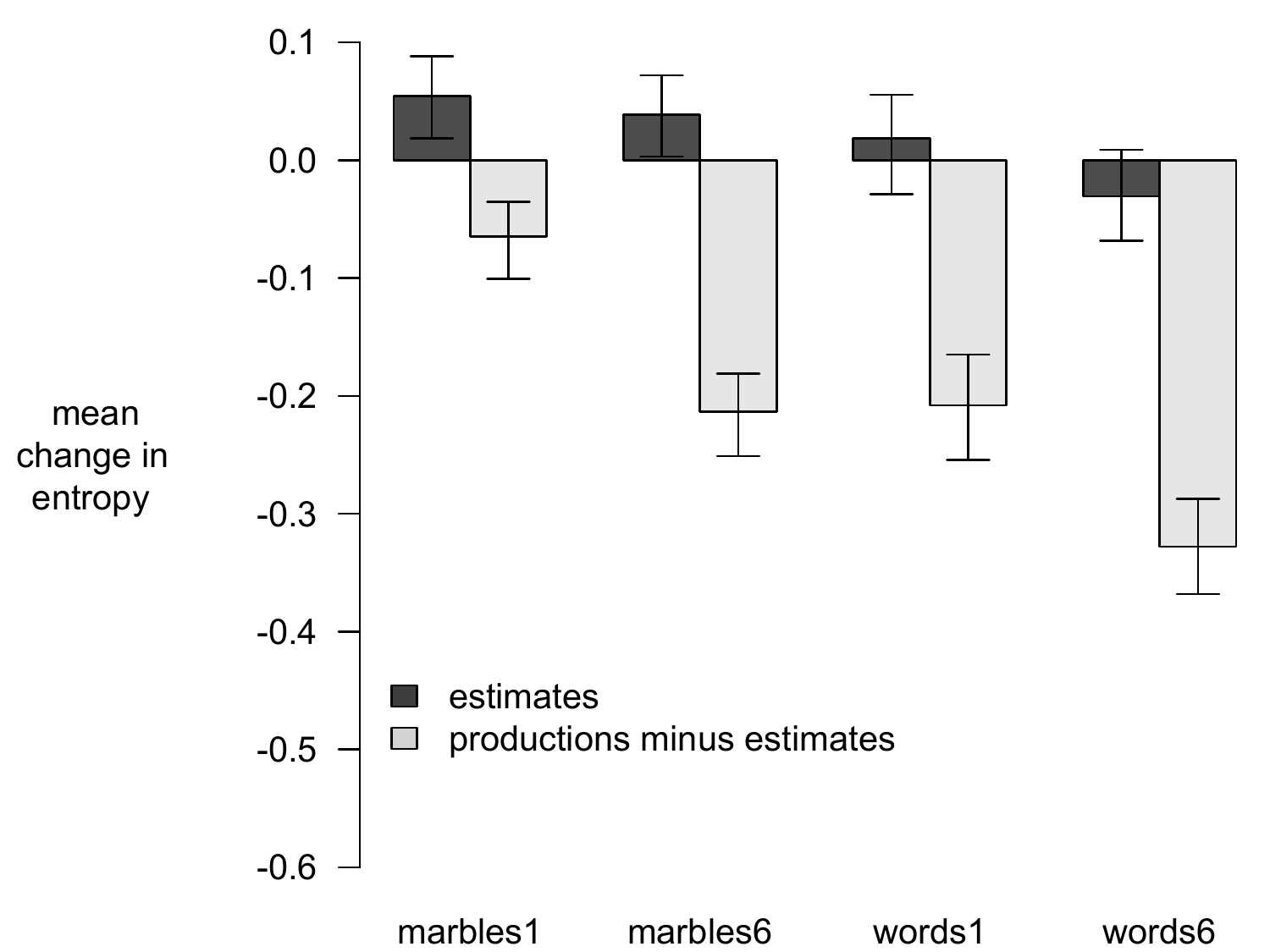}
\put(-164,176){*}
\put(-149.5,116){*}
\put(-110,82){***}
\put(-65.5,81){***}
\put(-21.5,55){***}
\caption{Production bias, not encoding bias, drives regularization.  
\emph{Dark grey:}  Average difference in regularity between the input ratios participants actually observed and their estimates of the underlying ratio that generated the input ratio.  A significant increase in entropy means that participants estimated the underlying ratio to be more variable than the input ratio, and a significant decrease means they estimated it to be more regular.  \emph{Light grey:}  Average difference between production ratio regularity and estimated ratio regularity.  Error bars indicate the 95\% confidence intervals computed with the bootstrap percentile method \citep{Efron1979}.}
\label{fig:4bars_Q1}
\end{figure}

\subsection{Regularization during encoding} \label{sec:encoding}

As discussed in the Introduction, regularization behavior is often explained as a result of general cognitive limitations on memory encoding and/or retrieval.
The high cognitive load manipulation in this experiment affected both the observation and production phases because both phases consisted of 60 interleaved trials.  
Therefore, the regularization behavior we observed could be due to encoding multiple frequencies under load (during the observation phase) and/or retrieving frequencies under load (during the production phase).  Furthermore, it is possible that linguistic domain may have a specific effect on the encoding of frequency information.
To determine whether encoding errors contribute to participants' regularization behavior in this experiment, we asked participants to estimate (using a slider) the underlying ratio that generated the marble draws or naming events they observed, per container or object (see Section \ref{sec:procedure}, last paragraph).  
If participants' estimates are not significantly different from the ratios they observed, then we can assume frequency encoding was unbiased.  This result would point to a production-side driver of regularization.

Figure \ref{fig:4bars_Q1} (dark grey bars) shows the average change in entropy between participants' estimates and the actual input ratios they observed.  The same linear mixed effects regression analysis described in Section \ref{sec:netreg} was applied to this data, using the change between input and estimate entropy as the dependent variable.  Only one condition, \emph{marbles1}, elicited a significant difference (of 0.05 bits) between the input ratios and estimates ($S.E. = 0.02, t(1152) = 2.29, p = 0.02$).  In this condition, participants estimated the generating ratio to be significantly \emph{more} variable than the ratio they had observed, indicating a slight encoding bias toward variability.  None of the conditions show any bias toward regularity in participants' estimates.
Effects of the experimental manipulations were assessed by the same procedure described in Section \ref{sec:manips_lmer}, using change between input and estimate entropy as the dependent variable.  The best-fit model contained a significant effect of domain ($\chi^2(4) = 11.735, p = 0.02$), cognitive load ($\chi^2(4) = 34.916, p < .001$), and input ratio ($\chi^2(4) = 562.04, p < .001$) (loglikelihood = $-72.558$, see Appendix Table \ref{tab:bestfit_encode}). One interaction was found to be a significant predictor of participants' estimates: cognitive load and input ratio ($\chi^2(1) = 27.916, p < .001$).  Interactions between domain and input ratio ($\chi^2(1) = 0.7554, p = 0.38$) and domain and cognitive load ($\chi^2(1) = 0.6741, p = 0.41$) were not significant.
Although the estimate data shows no bias toward regularity, the same factors that affected regularization behavior (cognitive load, domain, and input ratio) also affect participants' estimates.
Additionally, we find that the cognitive load manipulation resulted in noisier estimates ($F = 56.487, p < .001$, with Levene's test for homogeneity of variance), whereas the domain manipulation did not ($F = 0.4416, p = 0.51$).  This suggests that the high load condition was indeed more difficult than the low load condition and that the two domains were well-matched in terms of difficulty and stimuli complexity.

Figure \ref{fig:4bars_Q1} (light grey bars) shows the difference in entropy between the ratio participants produced and their estimate of that ratio, i.e. the extent to which their productions were more regular than their own estimate of their input data.  The same linear mixed effects regression analysis described in Section \ref{sec:netreg} was applied to this data, using the difference in entropy between the produced and estimated ratios as the dependent variable.  In all conditions, production ratios are significantly more regular than the estimates participants made (\emph{marbles1}: $S.E. = 0.03, t(1152) = -2.55, p = 0.01$; \emph{marbles6}: $S.E. = 0.03$,
\noindent
$t(1152) = -6.97, p < .001$; \emph{words1}: $S.E. = 0.03, t(1152) = -8.20, p < .001$; \emph{words6}: $S.E. = 0.03, t(1152) = -10.69$, 
\noindent\
$p < .001$).
This means that regularization occurs during the production phase and is likely to be involved in the retrieval and use of frequency information.  Interestingly, production-side regularization occurs in all four conditions, even in \emph{marbles1} where participants probability matched their productions to their inputs (effectively ``correcting" the variability bias in their estimates).  This suggests that regularity is broadly associated with frequency production behavior, even in cases that do not lead to overt regularization behavior.

In summary, raising cognitive load resulted in noisier encoding, however the noise was not biased in the direction of regularity.  Estimates in the linguistic domain were not biased toward regularity either.  It appears that the bulk of regularization occurs during the production-side of the experiment and is likely to involve processes of frequency retrieval and use.

\subsection{Individual differences in frequency learning strategy} \label{sec:individual}

The bimodal distributions over output ratios (refer back to Figure \ref{fig:2x2_6panels}) suggest individual differences in frequency learning strategies.  We break frequency learning behavior into three categories: \emph{regularizing}, \emph{probability matching}, and \emph{variabilizing}.  How many participants fall into each category?  And in the high load conditions, where participants respond to more than one item, how consistent are their responses with one strategy?

We define \emph{probability matching} as sampling from the input ratio, with replacement.  This leads to output ratios that are binomially distributed\footnote{Humans can probability match with variance that is significantly lower than binomial variance \citep[][pp.45-57]{Ferdinand2015}.  Therefore, the definition of probability matching used in this paper is a conservative one.} 
about the mean (where the mean equals the input ratio).  Although the single most likely output ratio a participant could sample is the set of input ratios itself, most probability matchers will sample a ratio that has higher or lower entropy than the input ratio.  We will classify participants who produced ratios within the 95\% confidence interval of sampling with replacement behavior as probability matchers.  We classify participants as \emph{variabilizers} if they produced ratios with significantly higher entropy than likely under probability matching behavior.  These could be participants who were attempting to produce a maximally variable set (all 5:5 ratios) or randomly selecting among the two choices on each production trial.  Likewise, we classify participants as \emph{regularizers} if they produced ratios with significantly lower entropy than likely under probability matching behavior.  It is important to note that a participant with a very weak bias for regularity or variability may consistently produce data that falls within the 95\% confidence range of probability matching.  However, we take a conservative approach by grouping individuals as regularizers or variabilizers only when probability matching has low probability.

\begin{figure*} [t] \centering
\includegraphics[width=180mm]{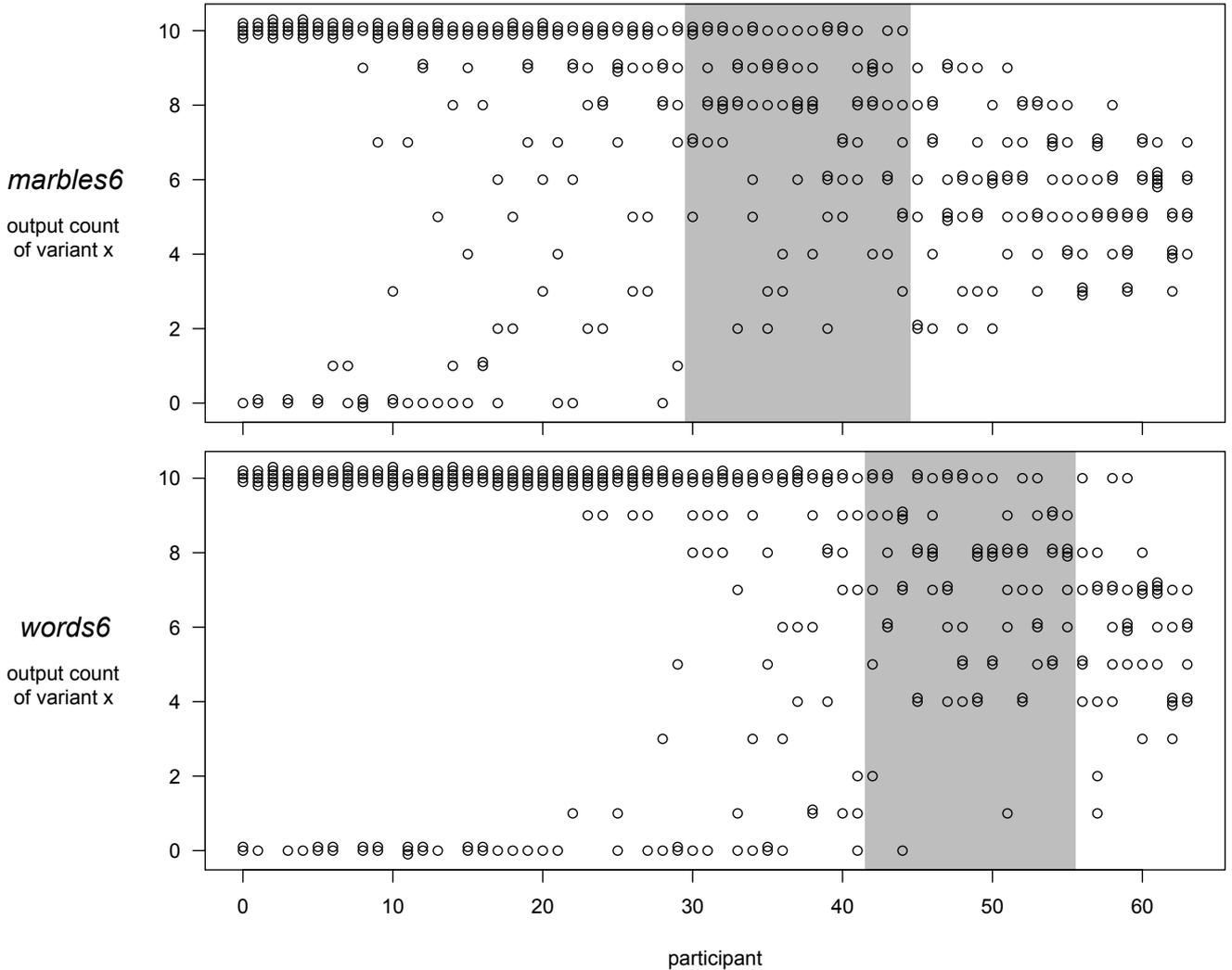}
\caption{Linguistic and non-linguistic stimuli evoke different frequency learning strategies.  Data are from the high cognitive load conditions \emph{marbles6} (top) and \emph{words6} (bottom). The x-axis shows participant number, sorted by their conditional entropy (low to high). The y-axis shows the frequency of the majority variant in the participant's output; each point represents performance on a single container/object, and there are therefore 6 points per participant.  The shaded region contains all participants classified as probability matchers. Participants to the left of the shaded region are classified as regularizers and participants to the right are classified as variabilizers.}
\label{fig:individual}
\end{figure*}

In the low load conditions, where participants only sample one ratio, the 95\% confidence intervals on output ratios were determined with the Clopper-Pearson exact method.\footnote{95\% confidence interval on probability matching per input ratio: \\ 5:5, $0.19 \leq x \leq 0.81$; 6:4, $0.26 \leq x \leq 0.88$; 7:3, $0.35 \leq x \leq 0.93$; 8:2, $0.44 \leq x \leq 0.97$; 9:1, $0.55 \leq x \leq0.99$; 10:0, $0.69 \leq x \leq 1$, where $x$ is the frequency of the majority variant.}
In the high cognitive load conditions, where participants sample a set of six ratios, we classify the set of ratios according to their conditional entropy $H(V|C)$ (refer back to Section \ref{sec:entropy}).  The 95\% confidence interval on conditional entropy for probability matching in this experimental setup is 0.43 to 0.75 bits (determined by $10^{5}$ runs of simulated probability matching behavior).  Participants who produced data with entropy in the range $0.43 \leq x \leq 0.75$ were classified as probability matchers, those who produced data in the range $0 \leq x < 0.43$ were classified as regularizers, and those who produced data in the range $0.75 < x \leq 1$ were classified as variabilizers.

\begin{table}[h]
\scalebox{0.88}{
\begin{tabular}{ | l | c | c | c | }
\hline
& regularizers & probability matchers & variabilizers \\
\hline
\emph{marbles1}  & 10 (5\%) & 173 (90\%) & 9 (5\%) \\
\hline
\emph{words1}  & 50 (26\%) & 139 (72\%) & 3 (2\%)  \\
\hline
\emph{marbles6}  & 30 (47\%) & 15 (23\%) & 19 (10\%) \\
\hline
\emph{words6}  & 42 (66\%) & 14 (22\%) & 8 (12\%) \\
\hline
\end{tabular}}
\caption{Participants classified by frequency learning strategy.  Percentages show how the strategies break down within each condition.}
\label{tab:individual}
\end{table}

Table \ref{tab:individual} shows the number of participants that fell into each frequency learning category, per condition.  All strategies are represented within each experimental condition.  There is a significant effect of cognitive load ($\chi^2(2) = 151.63, p < .001$) and domain ($\chi^2(2) =  31.49, p < .001$) on the distribution of frequency learning strategies, meaning that the experimental manipulations elicit different frequency learning strategies from participants.  Because fewer data points were collected from participants in the low load condition, probability matching behavior is not easily ruled out, hence the high number of participants classified as probability matchers in \emph{marbles1} and  \emph{words1}.  It is possible that the difference in dataset size between the low and high conditions is responsible for the significant effect of load.  The effect of domain, however, is reliably due to the experimental manipulation.  Therefore, the remainder of this section focuses on the high load data.

Figure \ref{fig:individual} shows the set of six output ratios that each participant produced in the high cognitive load conditions.  The sets are sorted by their entropy and the shaded box shows the sets that fell into the $0.43 \leq x \leq 0.75$ bit range (classified as probability matchers).  Participants to the left of the box are classified as regularizers and participants to the right are classified as variabilizers.  More regularizers were found in the linguistic domain, more variabilizers were found in the non-linguistic domain, and probability matchers seem equally likely to be found in either domain.
At the extreme left of the x-axis, we see the subset of regularizers, numbering 6 participants in \emph{marbles6} and 22 in \emph{words6}, who produced a maximally regular set (all 10:0 or 0:10, conditional entropy = 0 bits).  No participants produced a maximally variable set (all 5:5 ratios, conditional entropy = 1 bit).  Participants are more likely to maximally regularize in the linguistic condition ($\chi^2(1) = 10.2857, p = .001$).  Although some participants regularized with the majority variant exclusively, \emph{no} participants regularized with the minority variant exclusively.  Points in the 0-4 range on the the y-axis correspond to output ratios that contained a large number of minority variant productions (i.e. the majority variant had frequency of between 0 and 4).  Most participants regularized with 1-2 minority variants and 4-5 majority variants.

In summary, we found that all frequency learning strategies, \emph{regularizing}, \emph{probability matching}, and \emph{variabilizing}, are present in each condition and the use of linguistic stimuli causes more participants to consistently regularize.

%%%%%%%%%%%%%%%%%%%%%%%%%%%%%%%%%%%%%%%%%%%%%%%%%

\subsection{Primacy and recency effects on regularization} \label{sec:primacy}

Studies on regularization often find that participants regularize by over-producing or over-predicting the majority variant, and this serves as the standard definition of regularization \citep[e.g.][]{HudsonKam2005}.  However, many studies report some participants who regularize with the minority variant \citep[e.g.][]{HudsonKam2009,Reali2009, Smith2010a, Culbertson2012, Perfors2012, perfors2016adult}.  What causes some participants to regularize with the majority variant, and others to regularize with the minority variant?  In the previous section, we saw minority regularization is not due to individual differences in frequency learning behavior.  If minority regularization is not a feature of individuals, it may be a feature of the training data they received.

One possible data-driven explanation for minority regularization lies in the effects of a stimulus's primacy and recency on participant behavior.  In the observation phase, participants were presented with a randomly-ordered sequence of variants, such that the probability of any particular variant occurring at the beginning or end of the input sequence is proportional to its frequency in the sequence.  Therefore, some participants would have received minority variants toward the beginning and/or end of the sequence, whereas others would have not.
Many experiments on the serial recall of lexical items show that participants are better at recalling the first and last few items in a list of words \citep[e.g.][]{Deese1957,Murdock1962}.  This effect also extends to the learning of mappings between words and referents: \citet{Poepsel2014} found that when participants in a cross-situational learning task were confronted with several possible synonyms for an object, their confidence in a correct mapping was positively correlated with the primacy of that mapping in the observation phase.  Therefore, we investigated the effect of the minority variant's position in the input sequence on participants' tendency to regularize with the minority variant.

Unlike most research on primacy and recency (which present participants with a long list of unique stimuli), our input sequences only consist of two variants, presented several times each.  Therefore, we can quantify the strength of minority primacy as the imbalance of the variants across the input sequence.  To do this, we will use the notion of net torque.  In this analogy, we consider the input sequence to be a weightless lever of length 10 (the number of observation trials), we consider each minority variant to be a weight of one unit which is placed on the lever according to its observation trial number, and we assume the lever is balanced on a fulcrum at its center.  The sum of the distance of the weights located right of center minus the sum of the distance of the weights left of center is the net torque.  We will use the following standardization of net torque\footnote{Thanks to Andrew Berdahl for providing this solution.  The score was standardized in order to de-correlate net torque with input ratio.}, and refer to it as the primacy score:

\begin{equation}
primacy\ score = - \Bigg(\frac{ \sum\limits_{d=1}^N w_d(d - \frac{N+1}{2})}
{m(N-m)/2} \Bigg)
\end{equation}
\label{position_score}

\noindent
where $w$ is the sequence of weights and $d$ is the distance of that weight from the start of the sequence.  In the 5:5 input sequences, a random variant is coded as the ``minority" variant.  $N$ is the length of $w$ and $m$ is the total number of minority variants in the sequence.  Positive values mean that the minority variants occur more toward the beginning of the sequence and negative values mean they occur more toward the end of the sequence.  The maximum primacy score is 1 and the minimum is -1.  The average primacy score is 0 and is obtained when the sequence is balanced (i.e. minority variants are equally distributed early and late in the input sequence).  For example (where 1 indicates an occurrence of the minority variant in the input sequence), the primacy score of sequence 1110000000 is 1, 0000000001 is $-1$, 0101001000 is 0.33, 1000000001 is 0, and 0000110000 is 0.

\begin{table}[h]
\centering
\raggedright
\scalebox{0.78}{
\begin{tabular}{ | l | c | c | c | c | }
\hline
production sequences & \emph{marbles1} & \emph{words1} & \emph{marbles6} & \emph{words6} \\
\hline
total  & 192 & 192 & 384 & 384 \\
\hline
regularized  & 43 & 85 & 201 & 241 \\
\hline
regularized w/minority  & 16 (37\%) & 18 (21\%) & 53 (26\%) & 63 (26\%) \\
\hline
\end{tabular}}
\caption{Number of regularized production sequences per condition. Parentheses show the number of minority-regularized sequences as a percentage of all regularized sequences.}
\label{tab:reggers1}
\end{table}

Primacy analyses were restricted to the input sequences that participants regularized.  Table \ref{tab:reggers1} shows a breakdown of the number of regularized production sequences per experimental condition (i.e. all output sequences that had lower entropy than their corresponding input sequence).  Participants regularized a total of 570 input sequences.

\begin{figure} [t] 
\includegraphics[width=90mm]{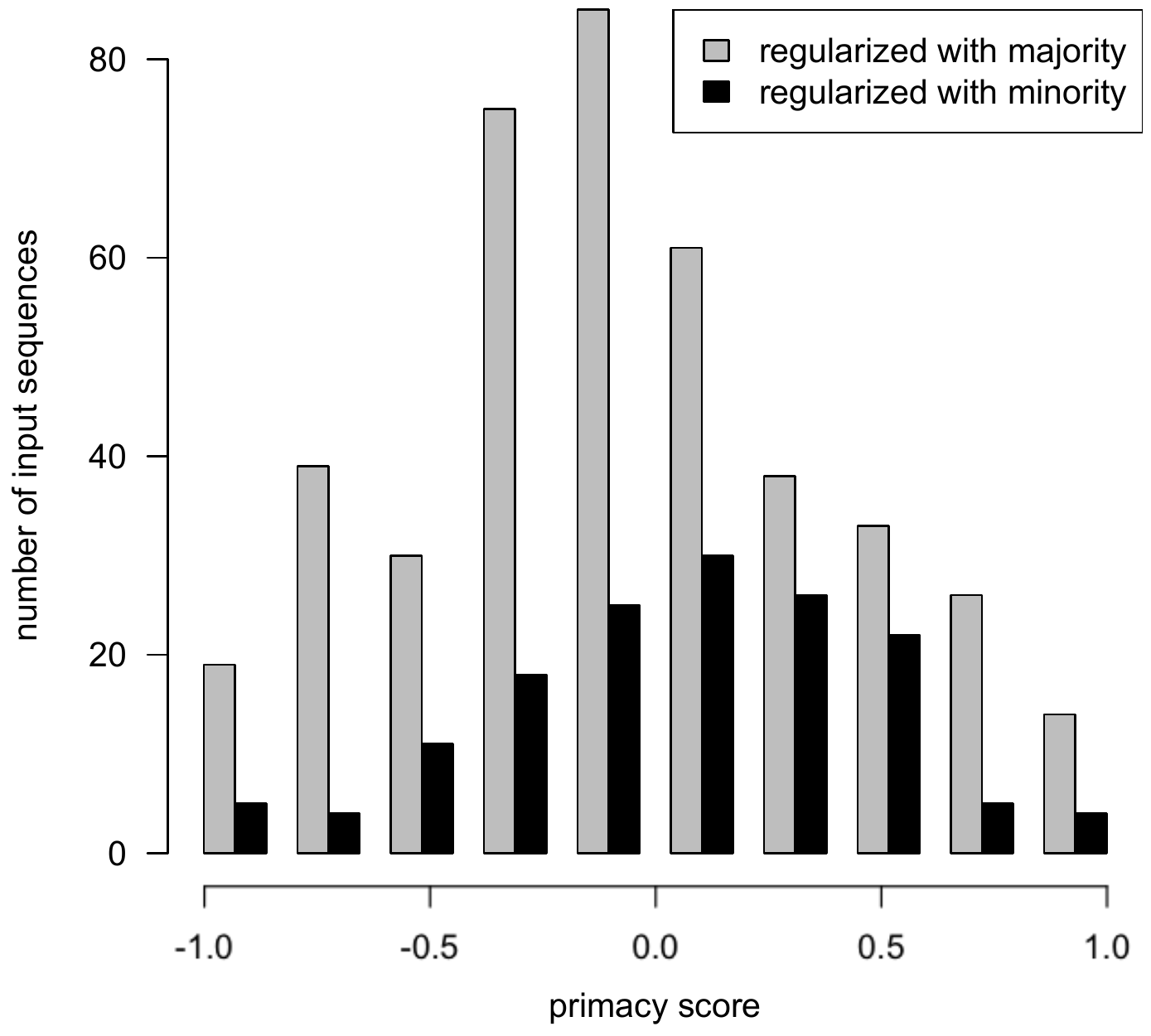}
\caption{Participants are more likely to regularize with the minority variant when they observe it toward the beginning of the input sequence.
The $x$-axis is the primacy of the minority variant in the input sequence, ranging from $-1$ (maximal recency) to 1 (maximal primacy).  Bars show the number of input sequences that were regularized by over-producing the minority variant (black) and by over-producing majority variant (grey).}
\label{fig:primacy}
\end{figure}

Figure \ref{fig:primacy} plots the primacy scores of the 420 sequences that were regularized with the majority variant (grey) and the 150 sequences that were regularized with the minority variant (black).  We constructed a logit mixed effects model of regularization type (majority or minority regularization) as a function of primacy score.  Participant was entered as a random effect (with random intercepts).  A likelihood ratio test was performed on this model and a reduced model which omits primacy score as a predictor.  We found a significant effect of primacy score on regularization type ($\chi^2(1)=6.4082, p = 0.01$).  On average, primacy score is 0.11 points higher ($\pm$ 0.04 standard errors) in sequences that were regularized with the minority.  This means that participants are more likely to regularize with the minority variant when they saw it toward the beginning of their input sequence (i.e. when minority variant primacy is high). 
However, minority regularization is not entirely explained by minority primacy.  As can be seen in Figure \ref{fig:primacy}, minority regularization was obtained across all primacy scores and even when the minority was maximally recent (left-most black bar).

In summary, we found that participants who saw the minority variant toward the beginning of the observation phase were more likely to regularize with the minority variant.  This helps explain some of the individual differences in regularization behavior, by grounding those differences in the properties of the data each participant observed.

%%%%%%%%%%%%%%%%%%%%%%%%%%%%%%%%%%%%%%%%%%%%%%%%%

\section{Predicting the evolution of regularity}

\begin{figure*} [t]
\includegraphics[width=183mm]{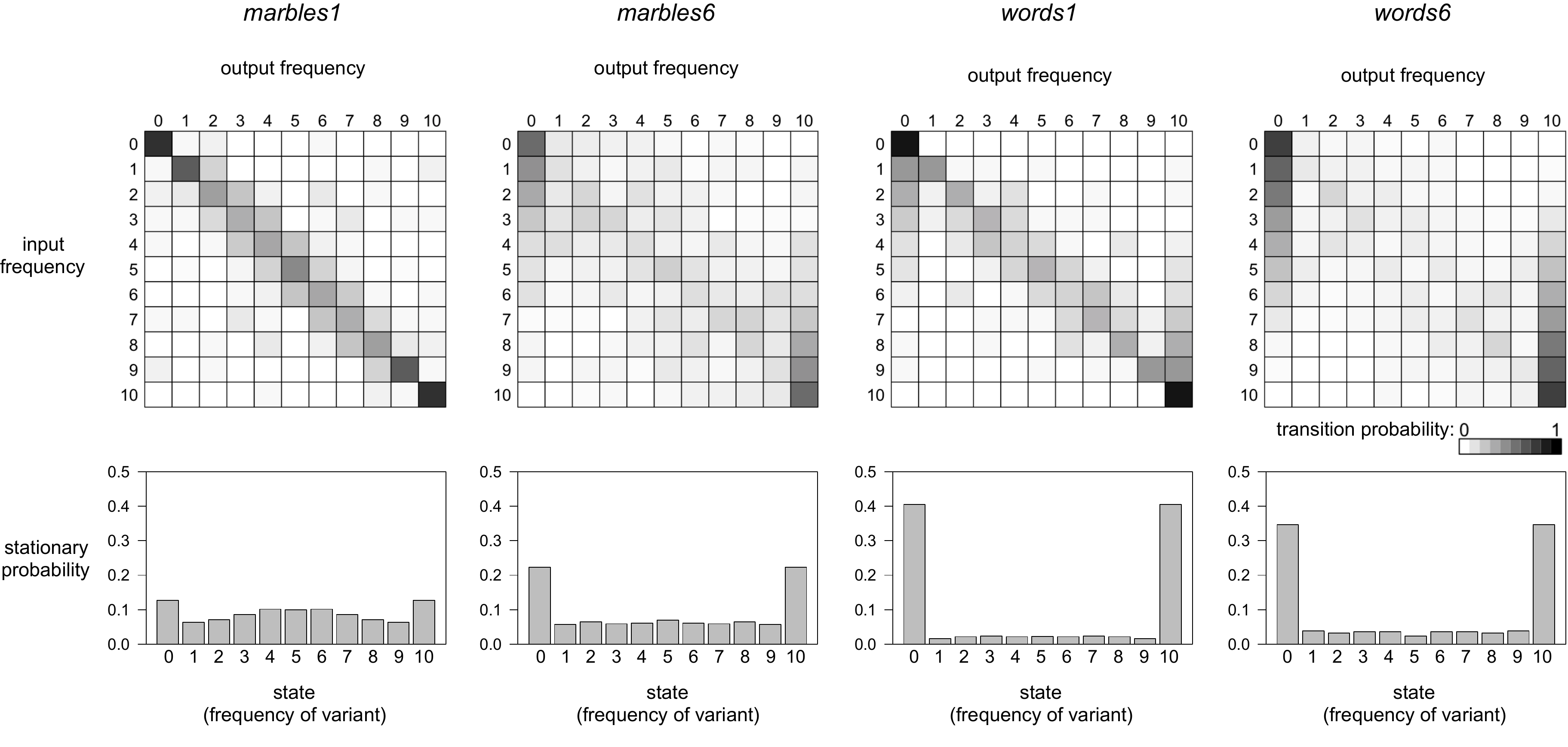}
\caption{The data from the experiment is used to predict the cultural evolution of regularization.
\emph{Top}:  Estimated transition matrices for each experimental condition contain the probabilities that a learner produces any given output ratio from any given input ratio (presented in terms of the frequency of variant $x$ in each input ratio \emph{x:y}).  The shading of the cells denote the transition probabilities between states.  Each row in the matrix corresponds to the distribution of output ratios produced in response to one input ratio (rows are the same distributions in Figure \ref{fig:2x2_6panels}, only smoothed).  For example, row 5 in the \emph{marbles1} transition matrix corresponds to the upper left panel of Figure \ref{fig:2x2_6panels}, and the probability of transitioning from $s_{t-1=5}$ to $s_{t=6}$ is equivalent to the (smoothed) proportion of participants that produced a 6:4 ratio when trained on a 5:5 ratio.  Likewise, rows 4 and 6 correspond to the 6:4 panel in Figure \ref{fig:2x2_6panels}, but this distribution is flipped in row 4 to display the results in terms of the minority variant.  \emph{Bottom}:  The stationary distribution shows the percentage of learners who will produce each output ratio, after the ratios have evolved for an arbitrarily large number of generations.  Each stationary distribution is the solution to the matrix above it.}
\label{fig:Qmatrices}
\end{figure*}

In the previous sections, we showed that learners regularize novel word frequencies due to domain-general and domain-specific constraints. This was accomplished by analyzing \emph{one cycle of learning}, which spans the perception, processing, and production of a set of variants.  Although this informs us about the relevant constraints that may underpin regularity in word learning, and even \emph{how much} regularity each constraint imposes on a given data set, it does not necessarily tell us how much regularity we will expect to see in a set of linguistic variants over time.  This is because languages are transmitted between generations of learners and are therefore subject to multiple learning cycles, where each individual has an opportunity to impose some amount of regularity on the language.

In this section, we address the complex relationship between regularization biases and the level of regularity found in culturally transmitted data.  In particular, we will focus on the evolution of regularity in the \emph{marbles6} and \emph{words1} conditions, because these two conditions elicited similar amounts of regularization behavior from two very different causes: domain-general and domain-specific constraints on frequency learning.  Would a data set which is culturally transmitted under conditions of only high cognitive load (as in \emph{marbles6}) or only linguistic framing (as in \emph{words1}) ultimately acquire the same amount of regularity?  

To answer this question, we will explore the dynamics of change in our existing data using an iterated learning model of cultural transmission \citep{Kirby2014iterated} in which the output of one learner serves as the input to another \citep[e.g.][]{Kirby2001,Brighton2002,Smith2003,Kirby2008,Reali2009,Smith2010a}.  Several cycles of iterated learning result in a walk over the complex landscape of constraints that shape the transmitted behavior, and several walks can be used to estimate this landscape and its likely evolutionary trajectories.  \citet{Griffiths2007} have shown that iterated learning is equivalent to a Markov process, which is a discrete-time random process over a sequence of values of a random variable, $v_{t=1}, v_{t=2}, ..., v_{t=n}$, such that the random variable is determined only by its most recent value \citep[p.535]{Papoulis1984}:

\begin{equation}
P(v_{t} | v_{t=1}, v_{t=2}, ..., v_{t-1}) = P(v_{t} | v_{t-1})
\label{eq:Q}
\end{equation}

\noindent
This describes a memoryless, time-invariant, process in which only the previous value ($v_{t-1}$) has an influence on the current value ($v_t$).  This is the case for iterated learning chains when learners only observe the behaviors of the previous generation.  All of the possible values of the random variable constitute the state space of this system.  A Markov process is fully specified by the probabilities with which each state will lead to every other state and these probabilities between states can be represented as a transition matrix,  \textbf{Q} \citep[p.3]{Norris2008}.  The probabilities in \textbf{Q} are the landscape over which a culturally transmitted dataset evolves.

In our experimental data, each state $s$ corresponds to one of the eleven possible ratios: $s_0, s_1, ..., s_{10} =$ \{0:10, 1:9, 2:8, 3:7, 4:6, 5:5, 6:4, 7:3, 8:2, 9:1, and 10:0\}, where $s_{t-1}$ is the input ratio and $s_t$ is the output ratio.  Our experiment was designed so that \textbf{Q} could be estimated for each of the four experimental conditions, by collecting data from participants in each of the eleven possible states.  Figure \ref{fig:Qmatrices} (top row) shows the estimated transition matrix from each experimental condition.  Each estimation consists of the raw data in that condition, smoothed with a small value $\epsilon = \textstyle\frac{1}{length(row)^2}$.  Each cell in the matrix, \textbf{Q}$_{ij}$, gives the transition probability from state $s_{i=t-1}$ to state $s_{j=t}$.

The transition matrices can be used to estimate the regularity of the data after an arbitrarily large number of learning cycles.  No matter what start state is used to initialize an iterated learning chain, an arbitrarily large number of iterations will converge to a stationary distribution, $\vec{s}$.  The stationary distribution is defined as $\vec{s}$\textbf{Q}$ = \vec{s}$, meaning that once the data take the form of the stationary distribution and serve as the input to $\textbf{Q}$, the output will be the same distribution and the subsequent generations of data will not change anymore.  The stationary distribution is a probability distribution over all states in the system, where each probability corresponds to the proportion of time the system will spend in each state, and can be solved for any matrix by decomposing the matrix into its eigenvalues and eigenvectors: $\vec{s}$ is proportional to the first eigenvector.  Figure \ref{fig:Qmatrices} (bottom row) shows the stationary distribution for each transition matrix.  From these distributions, we see that an arbitrarily long iterated learning chain will produce maximally regular (0:10 and 10:0) ratios approximately 40\% of the time when participants are learning about 12 marbles and six containers (\emph{marbles6}) and approximately 80\% of the time when participants are learning about two words and one object (\emph{words1}).  The difference between stationary distributions here means that the evolutionary dynamics of these two experimental conditions differ.

\begin{figure} [t] \centering
\includegraphics[width=90mm]{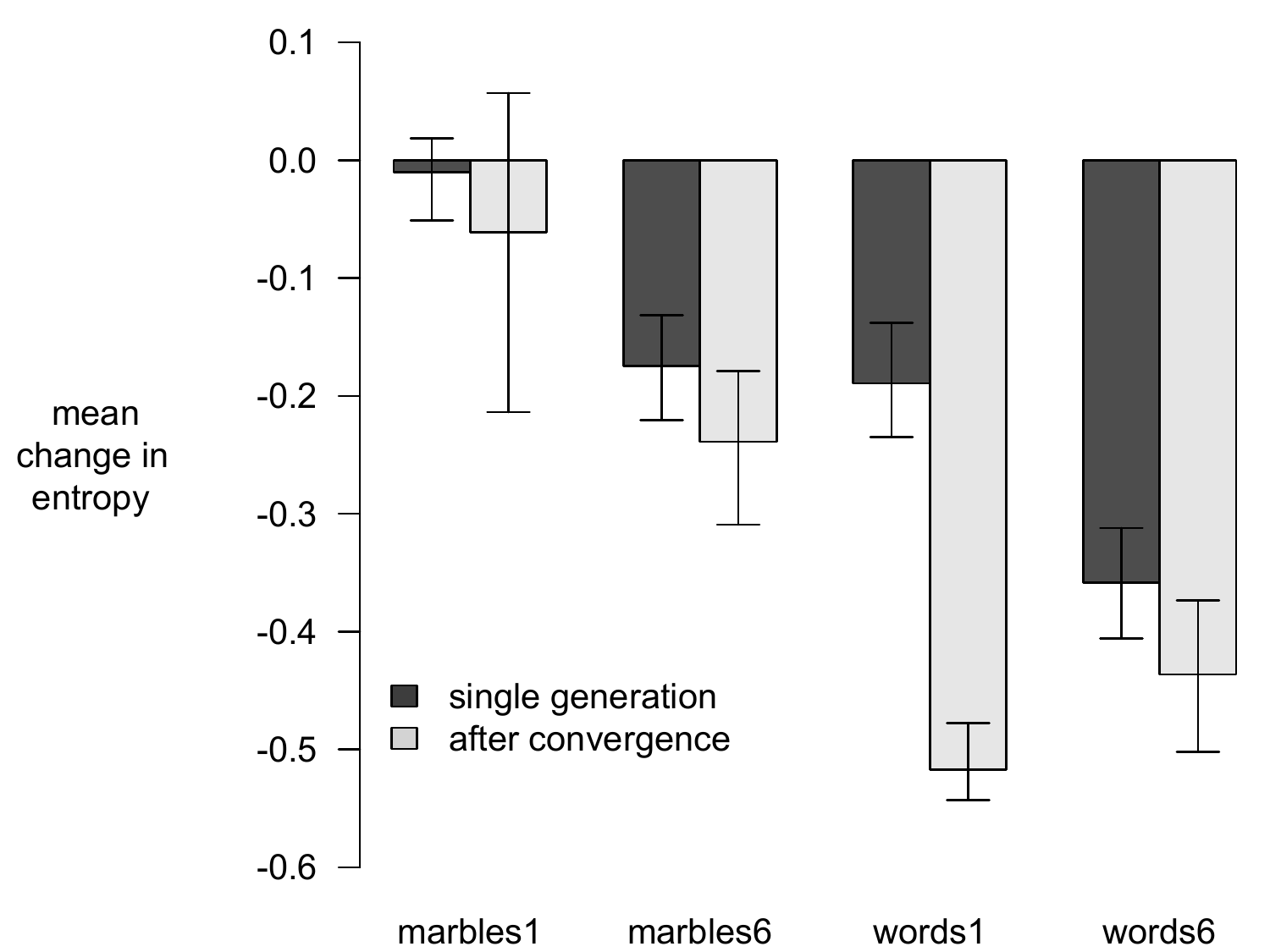}
\caption{Same learning biases lead to different degrees of regularization after many generations of cultural transmission.
\emph{Dark grey}:  Average change in entropy after one learning cycle (same data in Figure \ref{fig:4bars}, reprinted here for comparison).
\emph{Light grey}:  Average change in entropy of variants after convergence to the stationary distribution (i.e. after an infinite number of learning cycles).
Error bars indicate 95\% confidence intervals, computed by the bootstrap percentile method \citep{Efron1979} on 10,000 resamples of the transition matrix, where each matrix was solved for its stationary distribution and mean change in entropy.}
\label{fig:stat_reg}
\end{figure}

We calculate the level of regularity in the stationary distribution by multiplying the Shannon entropy of the ratio (defined by each state, $\vec{s}_i$) by the probability of observing that state, $p(\vec{s}_i)$.  The results are 0.61 bits of conditional entropy $H(V|C)$ in \emph{marbles1}, 0.43 bits in \emph{marbles6}, 0.16 bits in \emph{words1}, and 0.24 bits in \emph{words6}.  We compare these values to the results of the experiment (the average conditional entropy achieved after one learning cycle), which was 0.66 bits in \emph{marbles1}, 0.50 bits in \emph{marbles6}, 0.48 bits in \emph{words1}, and 0.32 bits in \emph{words6}.
Figure \ref{fig:stat_reg} plots these values in terms of entropy change: as the difference between the mean input entropy and the mean output entropy after one learning cycle (in dark grey) and after convergence to the stationary distribution (in light grey).  Here we see that, despite showing similar mean entropy change in the experiment, the regularization biases involved in \emph{marbles6} and \emph{words1} ultimately produce different levels of regularity via cultural transmission (inferred by the non-overlapping 95\% confidence intervals in stationary regularity between \emph{marbles6} and \emph{words1}).  This is due to the different distribution of probabilities within the transition matrices.  These probabilities constitute different landscapes that attract iterated learning chains into different regions of the state space.  One reason why the \emph{words1} data regularizes more than the other data sets, is that it has a markedly lower probability of transitioning out of the 10:0 and 0:10 states, trapping generations of learners in this highly regular region for longer amounts of time.

In summary, we have shown that the regularity elicited by two different constraints on frequency learning (the domain-general regularization biases involved in \emph{marbles6} and the domain-specific regularization biases involved in \emph{words1}) is similar in one generation of learners, but displays different evolutionary dynamics under simulated cultural transmission.  This finding has important implications for the relationship between learning biases and structure in language: it means that culturally transmitted systems, such as language, do not necessarily mirror the biases of its learners \citep[see][]{Kirby1999, kirby2004ug, smith2017language}.  Previously, we showed that cognitive load and linguistic domain are independent sources of  regularization in individual learners.  Looking at the data from individual learners, we may even infer that cognitive load and linguistic domain inject similar amounts of regularity into language.  However, the fact that \emph{words1} has higher stationary regularity than \emph{marbles6} means, at least in terms of the present data, that the amount of regularity we ultimately expect to find in a language is not simply predicted from a learner's biases.  Instead, the process of cultural transmission is an indispensable piece of the puzzle in explaining how learning biases shape languages.

%%%%%%%%%%%%%%%%%%%%%%%%%%%%%%%%%%%%%%%%%%%%%%%%%

\section{Discussion}

Regularity in language is rooted in the cognitive apparatus of its learners.  In this paper, we have shown that linguistic regularization behavior results from at least two, independent sources in cognition.  The first is domain-general and involves constraints on frequency learning when cognitive load is high.  The second is domain-specific and is triggered when the frequency learning task is framed with linguistic stimuli.

Cognitive load was manipulated by varying the number of stimuli in a frequency learning task.  When participants observed and produced for more stimuli, they regularized stimuli frequencies \emph{more} on average than when they were observing and producing for fewer stimuli.  This result held when stimuli were non-linguistic (marbles and containers) and when stimuli were linguistic (words and objects) and has previously been observed in separate non-linguistic and linguistic experiments \citep{Gardner1957,HudsonKam2009}.  We have shown, within the same experimental setting and for identical distributions of variation, that increasing cognitive load causes participants to regularize both non-linguistic and linguistic stimuli.  Furthermore, we have shown that participants regularize a similar amount of variation in both cases, eliminating 24.6\% of the variation in marbles conditioned on containers and 25.5\% of the variation in words conditioned on objects.  This similarity suggests that learners have general limits on the amount of variation they can process and reproduce, which are independent of the learning domain.  It is quite possible that cognitive load makes a fixed contribution to regularization behavior, however it remains to be seen whether this result holds over different learning domains and cognitive load manipulations.

One possible alternative explanation for the cognitive load effect on regularization behavior is the differing length of the two tasks.  Our design kept duration per stimulus constant across conditions, rather than total task duration, because it is unknown how stimuli presentation length affects regularization behavior.  However, it is possible that participants' attention was lower at the end of the high cognitive load tasks, causing them to over-produce the stimuli they saw early on in the training phase.  Given our finding that primacy affects minority regularization more than recency, this could be the case, although that effect was quite small.  Future research should address the effect of stimulus duration and presentation order on the degree to which participants regularize.

Domain was manipulated by varying the type of stimuli used in the frequency learning task.  When participants observed and produced mappings between words and objects, they regularized more than participants who observed and produced mappings between marbles and containers.  Participants appear to have a higher baseline regularization behavior when learning about linguistic stimuli: an additional 27\% of variation was regularized due to linguistic domain in each cognitive load condition (26.7\% in the low condition, 27.4\% in the high condition).  

The use of linguistic stimuli may trigger any number of domain-specific learning mechanisms or production strategies.  One possibility is that the stimuli manipulation changed participants' pragmatic inferences about the frequency learning task.  In an artificial language learning task, \cite{perfors2016adult} showed that participants regularize more when they believe that the variation in labels for objects can be the result of typos, suggesting that participants are more likely to maintain variation when they think it is meaningful.  It is possible that participants make different assumptions about the importance of variation in marbles versus words when they are required to demonstrate what they have learned to an experimenter.  However, it is not clear what these assumptions may be.  Another possibility is that the use of linguistic stimuli encourages participants to consider the communicative consequences of variation.  Participants in artificial language learning tasks regularize more when they are allowed to communicate with one another \citep{feher2016structural,smith2014eliminating} and even when they erroneously believe they are communicating with another participant \citep{feher2016structural}.  This suggests that participants strategically regularize variation in situations that are potentially communicative and may be the reason that regularization is observed in a wide range of language learning tasks, including the present study.

The use of non-linguistic stimuli may not fully put participants into a non-linguistic task framing: participants may be saying ``orange", ``blue", etc as they observe the marbles or verbalizing a rule such as ``there are more blue marbles in the jar".  If humans rely on language for solving complex problems, they may trigger linguistic representations when stimuli become more complex, regardless of the learning domain.  Following this logic, any increase in cognitive load could make any task more linguistic.  This would change the interpretation of our results, such that degree-differences in regularization behavior map on to degree-differences in the amount of linguistic representation involved.  Adopting this interpretation of our experiment, however, would require an overhaul of the definition of ``domain-general learning mechanisms" in statistical learning.

We also investigated the role of encoding errors on regularization behavior.  After the production phase, participants estimated the ratio of the variants associated with each container or object they had observed.  
We found that the same factors that affected production data also affected estimates (domain, cognitive load, and input frequency). %, but more variation was accurately encoded overall.
% We found that domain and cognitive load affect estimates:  more variation is encoded when cognitive load is low and stimuli are non-linguistic.  % TO DO
However, the estimates themselves were not significantly more regular than the input ratios that participants observed.  This suggests that participants had access to somewhat accurately encoded frequency information when making their estimates.  Because participants regularized their productions without showing a corresponding bias in estimates, this implies that the bulk of their regularization occurred during the production phase of the task.  This production-side interpretation is in line with the results of \citet{Chang2009}, who showed that adult participants regularize more when stimuli retrieval is made harder; \citet{Perfors2012}, who found adult participants do not regularize when encoding is made harder; and \citet{schwab2018regularization}, who show that children regularize during production despite demonstrated awareness of all word forms used during training.  This result also suggests that the Less-is-More hypothesis \citep{Newport1990}, which states that learners regularize because they fail to encode, store, or retrieve lower-frequency linguistic forms, applies more to retrieval and less to encoding.  However, it is possible that biased encoding could result from more complex mappings than those used in this experiment.\footnote{\citet{Vouloumanos2008} found that learners are able to encode and retrieve fine-grained differences in the statistics of low-frequency mappings between words and objects (which we calculate had a joint entropy of 4.41 bits), but failed to encode and retrieve fine-grained differences for a more complex stimuli set (with a joint entropy of 5.15 bits).  The joint entropy of our high cognitive load mappings, at 3.26 bits, is within \citet{Vouloumanos2008}'s demonstrated threshold for accurate frequency representation.}

An alternative explanation for the difference between the estimates and productions could be due to how these two types of data were elicited from participants.  It is likely the estimation question elicited more explicit knowledge about stimuli frequencies and the production task elicited more implicit knowledge (see \citealp{cleeremans1998implicit} for review).  If this is the case, it would mean that participants' explicit knowledge of observed frequencies is more accurate than their implicit knowledge and imply that regularization behavior is more closely associated with implicit knowledge retrieval.

This paper also explored the topic of minority regularization in depth.  We found that minority regularization is not the result of individual differences.  Although participants did differ in frequency learning strategies (we found regularizers, probability matchers, and variabilizers in all four conditions), most participants regularized with only one or two minority variants.  Therefore, we investigated differences in the randomized stimuli that each participant saw and found that participants are significantly more likely to regularize with the minority variant when it occurs toward the beginning of the observation sequence.  This primacy effect is in line with the results of \citet{Poepsel2014}, which showed that participants in a cross-situational learning task had higher confidence in the correctness of a mapping between words and referents when those items co-occurred early in the observation phase.  We also demonstrated how minority regularizers can confound regularization analyses which are based on the majority variant's change in frequency (Section \ref{sec:freq_analysis}) and argue that regularization should not be defined exclusively as ``overproduction of the highest-frequency or dominant form".  Alternative analyses that overcome this issue are \citet{Perfors2012,perfors2016adult}'s \emph{regularization index} and entropy-based analyses, as we use here (see Section \ref{sec:entropy}).  

There are several pros and a few cons to using entropy-based analyses.  Regularization occurs whenever learners increase the predictability of a linguistic system and therefore directly equates to a system's decrease in entropy.  Entropy measures allow us to quantify linguistic variation directly, in a mathematically-principled way based on predictability.  They also allow us to quantify all of the variation in a set of linguistic variants at once.  Analyses based on majority-variant frequency only tell us about changes to one or a subset of the variants in a language.  Overproduction of majority forms certainly can cause a language's entropy to drop, but regularity also can increase when minority forms are overproduced or when forms are maintained but conditioned on other linguistic contexts or meanings.  Entropy measures also force us to be explicit about what type of linguistic variation we are analyzing (i.e. the number of variants or their predictability in conditioning contexts) and allow direct comparison between different experiments (see footnote 5 as an example).  However, entropy and frequency analyses are sensitive to different aspects of linguistic data.  Entropy is better for quantifying regularization and positively identifying it, whereas frequency is better for detecting a population-level trend in over- or under-producing a particular variant.  For example, the frequency method used in Section \ref{sec:freq_analysis} did not capture the effect of cognitive load on frequency learning behavior, but it did capture an interesting domain difference that the entropy analysis missed: marble drawers overproduced the minority variant on average, whereas word learners did not.  These two methods also show differences in the classification of probability matching behavior: the entropy method identified \emph{marbles1} as consistent with probability matching behavior and the frequency method did not (because there is a significant bias toward the minority variant).  This raises important questions about the nature of probability matching: should it be defined as reproducing the same amount of variation (as the entropy measure captures) or reproducing the same amount of variation \emph{along with} the correct mapping of variation to stimuli (as the frequency measure captures)?

Overall, this paper explored how various cognitive constraints on frequency learning give rise to regularization behavior.  But what can detailed knowledge of these constraints tell us about the regularity of languages?  
One possibility is that the relationship between constraints on learning and structure of languages is straightforward, such that learning biases can be directly read off the typology of languages in the world \citep[e.g.][]{baker2002atoms} or the probability that a learner ends up speaking any given language can be read off the probability of the language in its prior \citep{Griffiths2007}.  Under other conditions, however, cultural transmission distorts the effects of learners' biases on the data they transmit, making it impossible to simply read learning biases off of language universals \citep{Kirby1999, kirby2004ug}.  Often, this distortion increases the effects of the bias over time, such that weak biases have strong effects on the structure of culturally transmitted data \citep[e.g.][]{Kalish2007,Kirby2007, Griffiths2008,Reali2009,Smith2010a, thompson2015transmission}.  However, the opposite can also occur: biases can have weaker effects or no effects at all \citep{smith2017language}.  This suggests that cultural transmission increases the complexity of the relationship between individual learning biases and the structure of language.  
By plugging the data obtained from our population of participants into a model of cultural transmission, we also found a complex relationship between regularization biases and regularity in culturally transmitted datasets.
Although participants produced similar amounts of regularity in response to the domain and load manipulations, the domain manipulation resulted in significantly higher regularity once the data was culturally transmitted.

Future research should explore whether similar effects of domain and demand on regularization behavior will be seen with child, rather than adult, learners. It is possible that child learners would show different relative contributions of domain general and domain specific biases than adults do.  Therefore, to the extent that language change is shaped specifically by biases in child acquisition \citep[see][for opposing views]{HudsonKam2005,slobin2014ontogenesis}, this may have implications for the role of language-specific and domain-general biases in shaping the evolution of linguistic regularity.

%%%%%%%%%%%%%%%%%%%%%%%%%%%%%%%%%%%%%%%%%%%%%%%%%

\section{Conclusion}

When learners observe and reproduce probabilistic variation, we find they regularize (reduce variation) when cognitive load is high and when stimuli are linguistic.  We conclude that linguistic regularization behavior is a co-product of domain-general and domain-specific biases on frequency learning and production.  Furthermore, we find that load and domain affect how participants encode frequency information.  However, encoded frequencies are not more regular than the data participants observed: the bulk of regularization occurs when participants produce data.  Finally, we show that the relative contributions of load and domain to the regularity in a set of linguistic variants can change when data are transmitted culturally.  In order to understand how various regularity biases create regularity in language, experiments that quantify learning biases need to be coupled with cultural transmission studies.

%%%%%%%%%%%%%%%%%%%%%%%%%%%%%%%%%%%%%%%%%%%%%%%%%

\section*{Acknowledgements}

This research was supported by the University of Edinburgh's College Studentship, the SORSAS award, and the Engineering and Physical Sciences Research Council.
The writeup was supported by the Omidyar Fellowship.
The reported experiment was conducted with the approval of the Linguistics and English Language Ethics Committee at the University of Edinburgh.
We thank Bill Thompson, Tom Griffiths, Florencia Reali, Simon DeDeo, Luke Maurits, and the members of the Centre for Language Evolution for their feedback on this project.  We thank Daniel Richardson for providing experimental stimuli and thank our reviewers, Amy Perfors and two anonymous, for their excellent comments.

\section*{Appendix}

%\phantom{invisible text}

\begin{table}[!htb]
\centering
\scalebox{0.87}{
%\scalebox{0.8}{
\begin{tabular}{ | l | c | c | c | }
\hline
 & estimate & S.E. & t statistic \\
\hline
Intercept  & 0.08567 & 0.04321 & $t(1152) = 1.983$ \\ 
\hline
Load (high) & 0.16715 & 0.04740 & $t(1152) = 3.527$ \\ 
\hline
Domain (words)  & $-0.11373$ & 0.04149 & $t(1152) = -2.742$ \\ 
\hline
Input entropy & $-0.14094$ & 0.05494 & $t(1152) = -2.565$ \\ 
\hline
Load * Input entropy & $-0.49568$ & 0.05622 & $t(1152) = -8.816$ \\ 
\hline
Domain * Input ent. & $-0.09968$ & 0.04596 & $t(1152) = -2.169$ \\ 
\hline
\end{tabular}}
\caption{The best-fit linear mixed effects model for change in entropy, see Section \ref{sec:manips_lmer}.}
\label{tab:bestfit1}
\end{table}

\begin{table}[!htb]
\centering
\scalebox{0.87}{
%\scalebox{0.8}{
\begin{tabular}{ | l | c | c | c | }
\hline
 & estimate & S.E. & t statistic \\
\hline
Intercept  & 0.11093 & 0.04804 & $t(1152) = 2.309$ \\ 
\hline
Domain (words)  & -0.12580 & 0.06794 & $t(1152) = -1.852$ \\ 
\hline
Input frequency & -0.21521 & 0.06217 & $t(1152) = -3.462$ \\ 
\hline
Domain * Input freq. & 0.22925 & 0.08792 & $t(1152) = 2.608$ \\ 
\hline
\end{tabular}}
\caption{The best-fit linear mixed effects model for change in frequency, see Section \ref{sec:freq_analysis}.}
\label{tab:bestfit_freq}
\end{table}

\begin{table}[!htb]
\centering
\scalebox{0.87}{
%\scalebox{0.8}{
\begin{tabular}{ | l | c | c | c | }
\hline
 & estimate & S.E. & t statistic \\
\hline
Intercept & $0.31193$ & 0.03127 & $t(1152) = 9.976$ \\ 
\hline
Load (high) & $0.13083$ & 0.03672 & $t(1152) = 3.563$ \\ 
\hline
Domain (words) & $-0.05145$ & 0.02001 & $t(1152) = -2.571$ \\ 
\hline
Input entropy & $-0.37040$ & 0.03901 & $t(1152) = -9.496$ \\ 
\hline
Load * Input ent. & $-0.24258$ & 0.04563 & $t(1152) = -5.316$ \\ 
\hline
\end{tabular}}
\caption{The best-fit linear mixed effects model for estimates, see Section \ref{sec:encoding}.}
\label{tab:bestfit_encode}
\end{table}

%%%%%%%%%%%%%%%%%%%%%%%%%%%%%%%%%%%%%%%%%%%%%%%%%

% \section*{References}

\bibliography{library}

\end{document}